\documentclass{article}

\usepackage{microtype}
\usepackage{graphicx}
\usepackage{subcaption}
\usepackage{booktabs} % for professional tables
\usepackage[normalem]{ulem}
\useunder{\uline}{\ul}{}

\usepackage{enumitem}

\usepackage{tabularx}
\newcolumntype{Y}{>{\centering\arraybackslash}X}

\usepackage{url}
\usepackage{amsmath}
\usepackage{bm}
\usepackage[table]{xcolor}
\usepackage[most]{tcolorbox}
\usepackage{booktabs}
\usepackage{multirow}
\usepackage{graphicx}
\usepackage{wrapfig}
\usepackage{subcaption} % 推荐使用subcaption宏包
\usepackage{arydshln}

\usepackage[preprint]{icml2026}

\usepackage{amsmath}
\usepackage{amssymb}
\usepackage{mathtools}
\usepackage{amsthm}

\usepackage{siunitx}    % 用于科学排版，特别是 S 列可以按小数点对齐数字
\usepackage{xfp} % 引入xfp包，用于进行精确的浮点数运算
\usepackage{etoolbox}     % 用于 \chg 命令中的条件判断
\usepackage{calc}         % \widthof 需要

\definecolor{upgreen}{rgb}{0.0, 0.5, 0.0} % 深绿色
\definecolor{downred}{rgb}{0.8, 0.0, 0.0} % 深红色

\newcommand{\formatChange}[1]{%
  {\scriptsize
    \ifdimcomp{#1 pt}{>}{0 pt}{% ---- 情况1：上升 ----
      ~({\color{upgreen}$\blacktriangle$#1})%
    }{%
      \ifdimcomp{#1 pt}{<}{0 pt}{% ---- 情况2：下降 ----
        ~({\color{downred}$\blacktriangledown$\fpeval{abs(#1)}})%
      }{% ---- 情况3：持平 ----
        ~(-)%
      }%
    }%
  }%
}

\newcommand{\chg}[2]{#1\formatChange{#2}}

\newcommand{\chgu}[2]{%
  #1%                                 <-- 1. 输出纯数字，供 siunitx 对齐
  {%                                  <-- 2. 开始“追加”下划线
    \llap{\underline{\phantom{#1}}}%   <-- 3. 在数字下方画一条等宽的下划线
  }%
  \formatChange{#2}%                  <-- 4. 附加变化值
}

\definecolor{lossgray}{gray}{0.5}      % 灰色：普通下降
\definecolor{worstred}{rgb}{0.8, 0.0, 0.0} % 红色：严重下降/最差结果
\definecolor{bestgreen}{rgb}{0.0, 0.6, 0.0} % 绿色：无损/最佳 (0.0)

\newcommand{\formatLossCore}[2]{%
  {\scriptsize%
    \ifdimcomp{#1 pt}{>}{0 pt}{% ---- 情况1：上升 (异常) ----
      ~({\color{worstred}$\blacktriangle$\num[round-mode=places, round-precision=1]{#1}})%
    }{%
      \ifdimcomp{#1 pt}{<}{0 pt}{% ---- 情况2：下降 ----
        ~({\color{#2}$\blacktriangledown$\num[round-mode=places, round-precision=1]{\fpeval{abs(#1)}}})%
      }{% ---- 情况3：持平 (0.0) ----
        ~({\color{bestgreen}$\blacktriangledown$\textbf{0.0}})%
      }%
    }%
  }%
}

\newcommand{\chgLoss}[2]{#1\formatLossCore{#2}{lossgray}}

\newcommand{\chgLossr}[2]{#1\formatLossCore{#2}{worstred}}

\usepackage[capitalize,noabbrev]{cleveref}

\theoremstyle{plain}

\theoremstyle{definition}

\theoremstyle{remark}

\usepackage[textsize=tiny]{todonotes}

\icmltitlerunning{From Absolute to Relative: Rethinking Reward Shaping in Group-Based Reinforcement Learning}

\begin{document}

% TITLES
% From Scalars to Rankings: Stabilizing Group Relative Policy Optimization with Intra-Prompt Comparisons
% Ranking is Reward: Intra-Group Preference Ranking for Group Relative Policy Optimization
% From Scalars to Rankings: Rethinking Reward Shaping in Group-Based Reinforcement Learning
% From Absolute to Relative: xxxxx

% Methods
% RLCR (Reinforcement Learning with Comparative Rewards)
% RLOR (Reinforcement Learning with Ordinal Rewards)

% RLGR (Reinforcement Learning with Group Rewards)
% RLRR (Reinforcement Learning with Ranking Rewards)

\twocolumn[
  \icmltitle{From Absolute to Relative: Rethinking Reward Shaping \\ in Group-Based Reinforcement Learning}

  % It is OKAY to include author information, even for blind submissions: the
  % style file will automatically remove it for you unless you've provided
  % the [accepted] option to the icml2026 package.

  % List of affiliations: The first argument should be a (short) identifier you
  % will use later to specify author affiliations Academic affiliations
  % should list Department, University, City, Region, Country Industry
  % affiliations should list Company, City, Region, Country

  % You can specify symbols, otherwise they are numbered in order. Ideally, you
  % should not use this facility. Affiliations will be numbered in order of
  % appearance and this is the preferred way.
  \icmlsetsymbol{equal}{*}
  \icmlsetsymbol{intern}{$\dagger$}

  \begin{icmlauthorlist}
    \icmlauthor{Wenzhe Niu}{equal,intern,tju,mt}
    \icmlauthor{Wei He}{equal,mt}
    \icmlauthor{Zongxia Xie}{tju}
    \icmlauthor{Jinpeng Ou}{mt}
    \icmlauthor{Huichuan Fan}{mt}
    \icmlauthor{Yuchen Ge}{mt}
    \icmlauthor{Yanru Sun}{tju}
    \icmlauthor{Ziyin Wang}{tju}
    \icmlauthor{Yizhao Sun}{mt}
    \icmlauthor{Chengshun Shi}{mt}
    \icmlauthor{Jiuchong Gao}{mt}
    \icmlauthor{Jinghua Hao}{mt}
    \icmlauthor{Renqing He}{mt}
  \end{icmlauthorlist}

  % \icmlaffiliation{yyy}{Department of XXX, University of YYY, Location, Country}
  \icmlaffiliation{tju}{Tianjin University, Tianjin, China}
  \icmlaffiliation{mt}{Meituan, Beijing, China}
  % \icmlaffiliation{pku}{Peking University, Beijing, China}    \icmlaffiliation{usc}{University of Southern California, Los Angeles, USA}
  
 % If you have any questions, feel free to contact (niuwenzhe@tju.edu.cn)
  \icmlcorrespondingauthor{Zongxia Xie}{caddiexie@hotmail.com}
  % \icmlcorrespondingauthor{}{}

  % You may provide any keywords that you find helpful for describing your
  % paper; these are used to populate the "keywords" metadata in the PDF but
  % will not be shown in the document
  \icmlkeywords{Large Language Model, Reinforcement Learning}

  \vskip 0.3in
]

% this must go after the closing bracket ] following \twocolumn[ ...

% This command actually creates the footnote in the first column listing the
% affiliations and the copyright notice. The command takes one argument, which
% is text to display at the start of the footnote. The \icmlEqualContribution
% command is standard text for equal contribution. Remove it (just {}) if you
% do not need this facility.

% Use ONE of the following lines. DO NOT remove the command.
% If you have no special notice, KEEP empty braces:
% \printAffiliationsAndNotice{}  % no special notice (required even if empty)
% Or, if applicable, use the standard equal contribution text:
\printAffiliationsAndNotice{\icmlEqualContribution 
$^\dagger$Work done during internship at Meituan.
}
% \printAffiliationsAndNotice{\icmlEqualContribution}

\begin{abstract}

Reinforcement learning has become a cornerstone for enhancing the reasoning capabilities of Large Language Models, where group-based approaches such as GRPO have emerged as efficient paradigms that optimize policies by leveraging intra-group performance differences. However, these methods typically rely on absolute numerical rewards, introducing intrinsic limitations. In verifiable tasks, identical group evaluations often result in sparse supervision, while in open-ended scenarios, the score range instability of reward models undermines advantage estimation based on group means. To address these limitations, we propose \textbf{Reinforcement Learning with Relative Rewards (RLRR)}, a framework that shifts reward shaping from absolute scoring to relative ranking. Complementing this framework, we introduce the \textbf{Ranking Reward Model}, a listwise preference model tailored for group-based optimization to directly generate relative rankings. By transforming raw evaluations into robust relative signals, RLRR effectively mitigates signal sparsity and reward instability. Experimental results demonstrate that RLRR yields consistent performance improvements over standard group-based baselines across reasoning benchmarks and open-ended generation tasks. 
\end{abstract}

\section{Introduction}

\begin{figure*}[!t]
  \centering
  \begin{subfigure}[t]{0.275\textwidth}
    \centering
    \includegraphics[width=\linewidth]{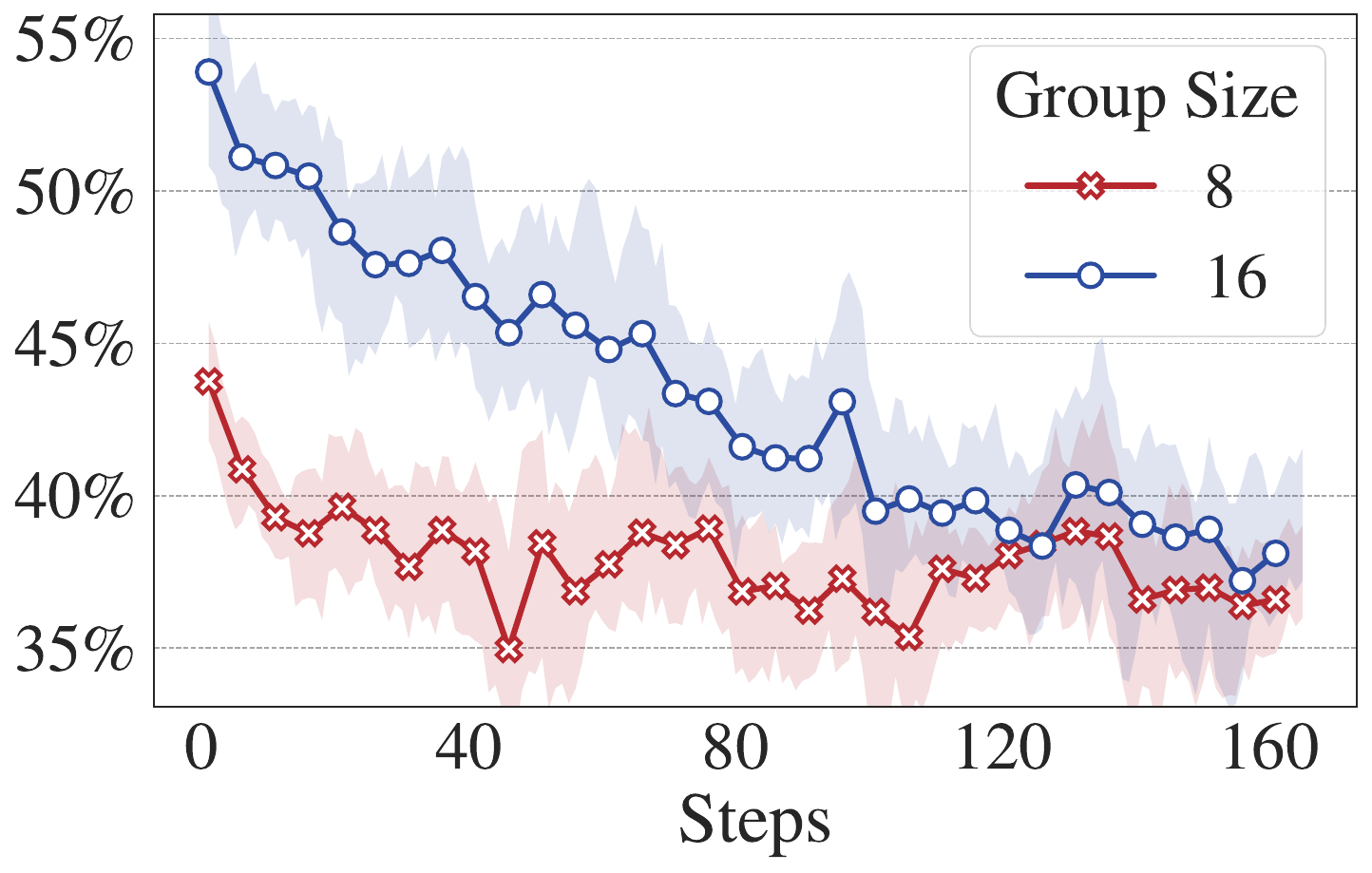}
    \caption[R]{\small Ratio of Effective Prompts}
    \label{fig:effective_prompts}
  \end{subfigure}
  \begin{subfigure}[t]{0.27\textwidth}
    \centering
    \includegraphics[width=\linewidth]{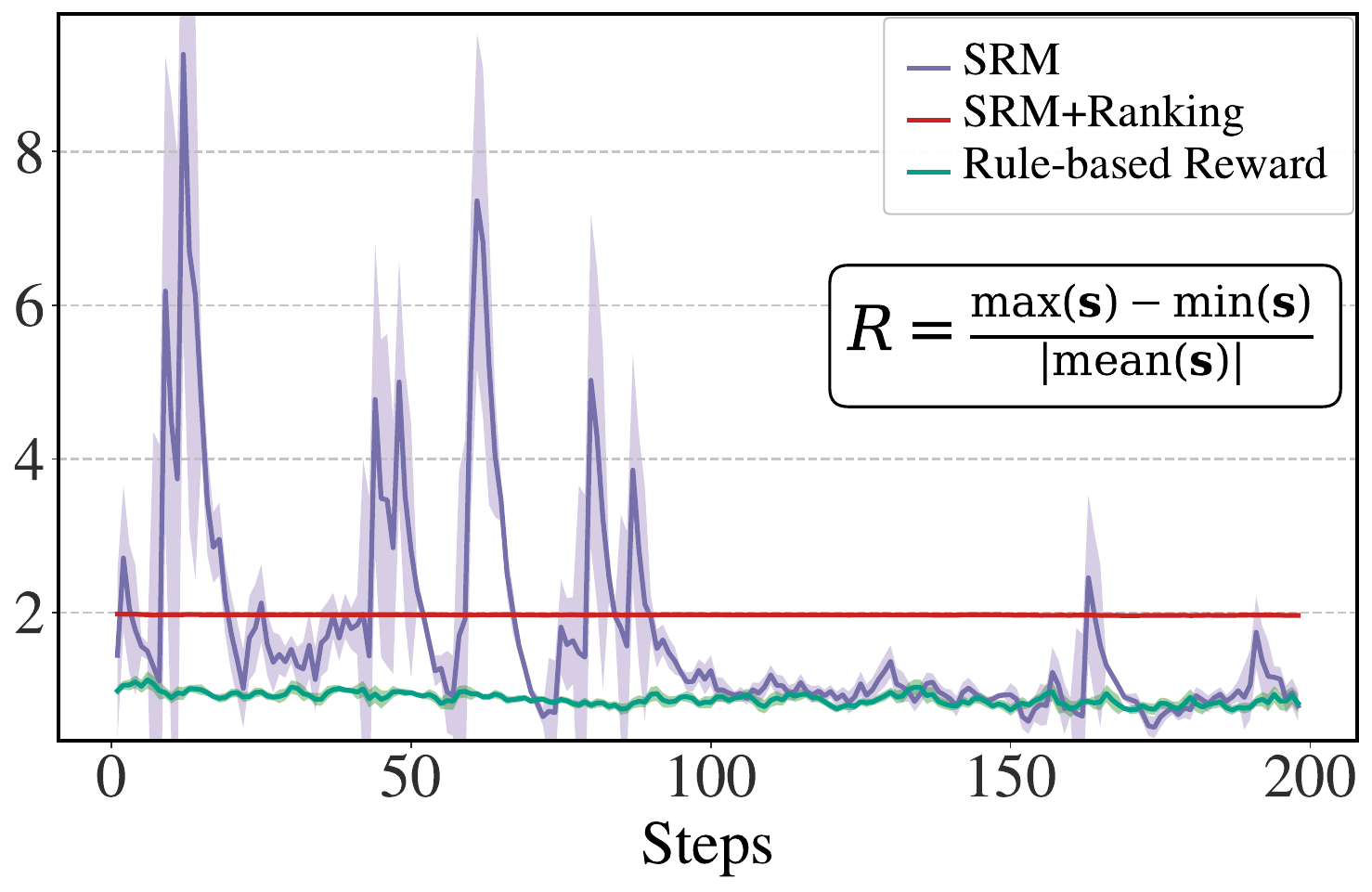}
    \caption[R]{\small Relative Range Dynamics}
    \label{fig:rm_stable}
  \end{subfigure}
  \begin{subfigure}[t]{0.43\textwidth}
    \centering
    \includegraphics[width=\linewidth]{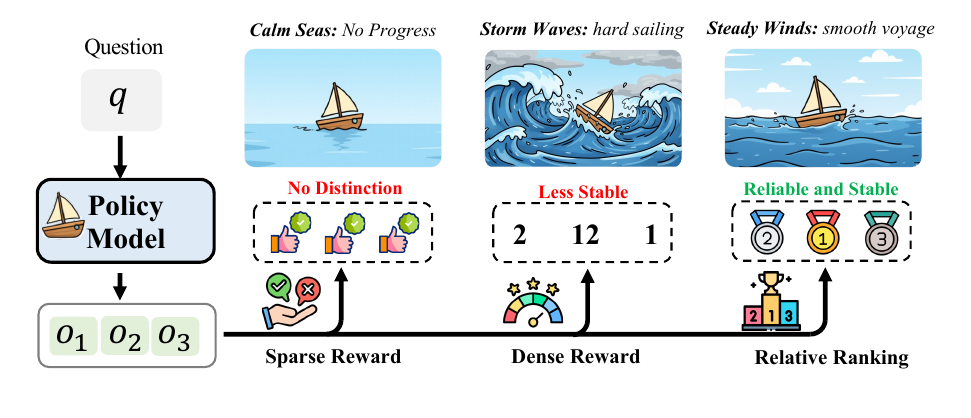}
    \caption[R]{\small Comparison of Different Rewards}
    \label{fig:sea}
  \end{subfigure}
  \caption{Comparative analysis of reward formulations. (a) Sparse rewards cause zero intra-group variance, reducing data utilization efficiency during GRPO training. (b) Scalar reward models exhibit higher numerical dispersion, compromising stability in intra-group advantage estimation. (c) RLRR utilizes relative ranking to sustain stable advantages through consistent gradient signals.}
  \label{fig:reward_compare}
  \vspace{-12pt}
\end{figure*}

In recent years, reinforcement learning (RL) has driven significant advances in natural language processing (NLP), reshaping the reasoning paradigms of large language models (LLMs). Through large-scale RL training, models such as DeepSeek-R1 \citep{guo2025deepseek} and OpenAI O1 \citep{jaech2024openai_o1} have demonstrated sophisticated reasoning abilities including self-verification and iterative refinement, which substantially improve their performance on challenging mathematical and programming tasks. Building on this progress, Group Relative Policy Optimization (GRPO) \citep{shao2024deepseekmath} has emerged as a key method for scaling LLMs during testing. By introducing an intra-group relative evaluation mechanism, GRPO reduces the bias of value function estimation and alleviates the heavy memory requirements associated with traditional Proximal Policy Optimization (PPO) \citep{schulman2017proximal-ppo}, providing a more efficient and robust training paradigm for the next generation of LLMs.

Reward modeling constitutes a central component in Reinforcement Learning, yet the reliance on absolute signal values creates a structural bottleneck for group-based optimization methods such as GRPO. Since these algorithms depend fundamentally on within-group difference to estimate advantages, absolute scoring mechanisms often fail to provide stable and continuous learning signals. In scenarios utilizing rule-based verifiers, the feedback is typically sparse and binary. As the policy improves during training, an increasing proportion of prompt groups reach a state of unanimous correctness or failure. We designate training prompts that yield mixed outcomes within a sampled group as \emph{effective samples}, as these are the only instances capable of inducing non-zero reward variance and meaningful policy gradients. However, as illustrated in \Cref{fig:effective_prompts}, the fraction of effective samples declines significantly to below 40\% in later training stages, implying that the majority of computational cost is expended on groups that contribute negligible optimization signals. 
While employing Scalar Reward Models (SRMs) can mitigate this sparsity by providing dense feedback, the unbounded nature of SRM scores implies that this approach introduces numerical instability. As shown in \Cref{fig:rm_stable}, this wide dynamic range undermines the stability of intra-group advantage estimation, where the absolute scale of rewards acts as a confounding factor. Consequently, current absolute reward paradigms either suffer from vanishing gradients due to outcome homogeneity or unstable updates due to score sensitivity. We provide a detailed analysis of the impact of the SRMs' unboundedness on group-based reinforcement learning in Appendix~\ref{app:theo_analysis}.

To address these limitations, we propose \textbf{Reinforcement Learning with Relative Rewards (RLRR)}, a framework designed to transition the reward shaping paradigm from absolute valuation toward relative ordering. By integrating intra-group ranking information directly into the advantage computation, RLRR effectively mitigates the dependency on the absolute magnitude of reward signals. We devise distinct relative reward mechanisms tailored to the availability of ground truth: for tasks equipped with rule-based verifiers, RLRR synergizes ranking signals with rule-based feedback; conversely, for tasks lacking deterministic verification, the framework relies exclusively on relative ranking. To facilitate this approach, we introduce the \textbf{Ranking Reward Model (Ranking RM)}, which is trained to discriminate the relative quality of multiple input samples. This design aligns with the inherent group-based comparison mechanism of GRPO. By employing the Ranking RM within the RLRR framework, our method is able to extract valid learning signals from sample groups that yield homogeneous outcomes under absolute scoring. This capability allows the model to continue learning from conventionally ineffective samples, thereby significantly enhancing reasoning quality and maximizing the utilization of rollout data. We rigorously validate the effectiveness of our methodology through extensive experiments across diverse settings, demonstrating substantial improvements over existing approaches.
Our contributions are summarized as follows:
\begin{itemize}[noitemsep,topsep=3pt]
    \item We propose RLRR, a framework that integrates relative rewards into group-based optimization to resolve signal sparsity in verifiable tasks and score instability in open-ended generation.
    \item We introduce the Ranking Reward Model, a listwise preference model that generates direct relative rankings, offering robust guidance immune to absolute score fluctuations.
    \item Extensive experiments validate the superiority of the relative reward paradigm, demonstrating consistent performance gains over absolute-scoring baselines across diverse benchmarks.
\end{itemize}

\section{Related Work}

\textbf{Inference-Time Scaling for LLMs.}
Inference-time scaling complements training efforts, with research focusing on sampling and reward model aggregation~\citep{brown2024large,snell2025scaling,wu2024inference}. A key approach is Reinforcement Learning with Verifiable Reward (RLVR), which improves reasoning by using external verifiers for reward signals instead of model-generated scores~\citep{zeng2025simplerl}. Methods like PPO~\citep{schulman2017proximal-ppo} and GRPO~\citep{shao2024deepseekmath} are commonly used for policy optimization, driving further RL advancements in reasoning tasks~\citep{kazemnejad2024vineppo, yuan2025vc_ppo}. Notable innovations include DAPO, which filters zero-variance prompts~\citep{yu2025dapo}, and GRESO, which uses probabilistic pre-filtering~\citep{zheng2025act_greso}. While both improve data efficiency, DAPO incurs computational overhead, and GRESO may discard useful learning opportunities due to its simplistic reward structure.

\textbf{Reward Models.}
Reward models (RMs) are pivotal in RL, especially for aligning LLMs and scaling inference. Designed to capture human preferences, RMs complement rule-based rewards~\citep{christiano2017deep, ouyang2022training}. Mainstream RMs typically function as discriminative classifiers, providing scalar rewards to rank responses~\citep{cai2024internlm2,liu2025skywork,lou2024uncertainty}. Other methods harness LLMs as judges, offering preference scores or critiques on generated content ~\citep{zheng2023llm}. Approaches like Direct Preference Optimization (DPO) eliminate the need for explicit RMs, instead directly optimizing policies from preference pairs ~\citep{rafailov2023direct}. Despite their advantages, RMs face challenges, such as the high cost of preference data, biases, and the risk of reward hacking ~\citep{gao2023scaling, skalse2022defining}.

\section{Preliminaries}

To optimize the LLM policy, GRPO~\citep{shao2024deepseekmath} introduces an alternative RL algorithm, which is a memory-efficient variant of PPO~\citep{schulman2017proximal-ppo}. A notable feature of GRPO is that it typically operates without a learned value function. Instead, for a given prompt $\bm{p}$, the current policy generates a group of $G$ responses $\{\bm{o}_1, \ldots, \bm{o}_G\}$. The rewards $\{s_1, \ldots, s_G\}$ for these responses are then used to compute the relative advantage for each response:
\begin{equation}
	\label{eq:advantage}
	\hat{A}_{k} = \frac{s_k - \text{mean}(\{s_k|k=1,2,\ldots,G\})}{F_{\text{norm}}}.
\end{equation}
Here, \( F_{\text{norm}} \) serves as an optional normalization factor. In the standard GRPO implementation, \( F_{\text{norm}} \) is defined as \( \text{std}(\{s_k|k=1,\ldots,G\}) \). In contrast, alternative implementations in RLVR fix the normalization factor to unity so that \( F_{\text{norm}} = 1 \) \citep{liu2025understanding,chu2025gpg}.

GRPO then maximizes a clipped surrogate objective function to ensure stable updates. Let $\pi_{\theta_{\text{old}}}$ represent the policy before the update. For each token $o_{k,t}$ in a trajectory $\bm{o}_k$ (from state $s_{t}$), the importance sampling ratio is defined as \( \rho_{k,t}(\theta) = \frac{\pi_{\theta}(o_{k,t}|s_{t})}{\pi_{\theta_{\text{old}}}(o_{k,t}|s_{t})} \). The objective is then given by:
\begin{equation}
\begin{aligned}
\mathcal{J}_\text{GRPO}(\theta) = \frac{1}{G} \sum_{k=1}^{G} & \frac{1}{|\bm{o}_k|} \sum_{t=1}^{|\bm{o}_k|}
\min\Bigl( \rho_{k,t}(\theta) \cdot \hat{A}_{k}, \\
& \text{clip}\bigl(\rho_{k,t}(\theta), 1-\epsilon, 1+\epsilon\bigr) \cdot \hat{A}_{k} \Bigr),
\end{aligned}
\label{eq:grpo_obj}
\end{equation}
where $\epsilon$ is a small hyperparameter defining the clipping range. This mechanism ensures that the LLM policy is updated while maintaining stable gradient constraints.

\begin{figure*}[!t]
  \centering
  \includegraphics[width=\linewidth]{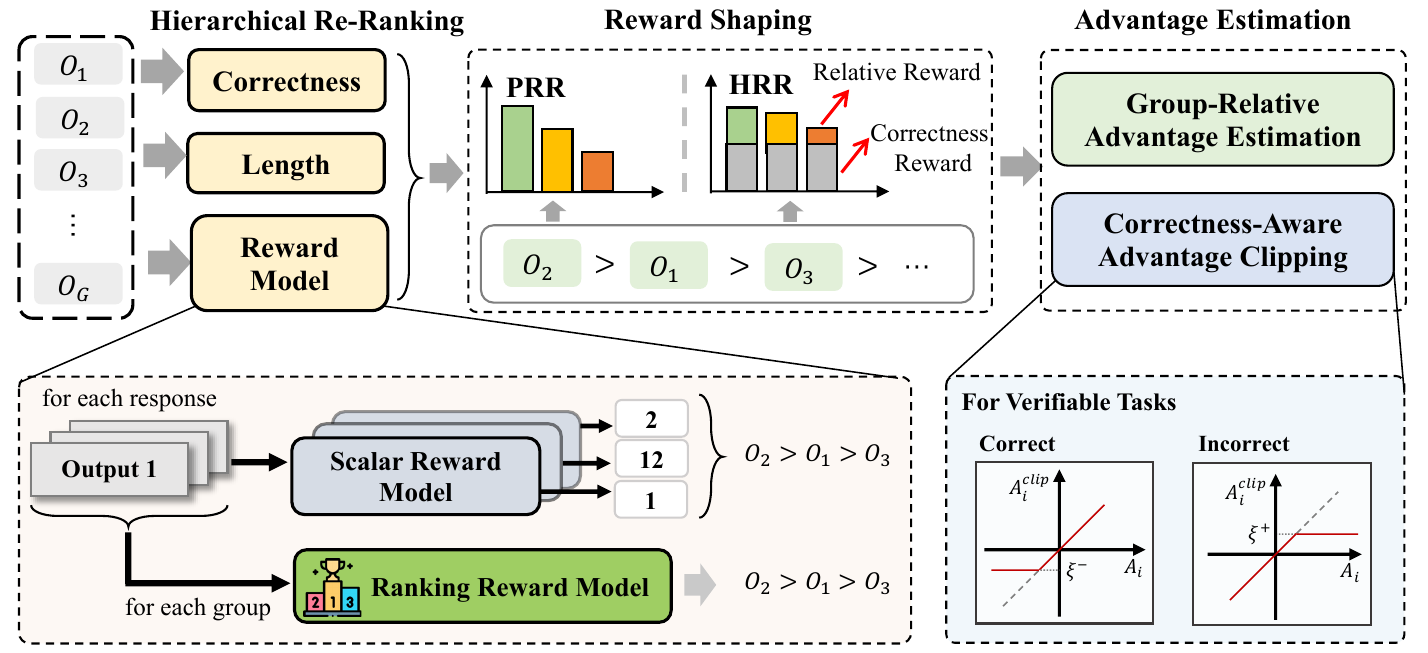} % 调整宽度
\caption{Overview of RLRR. The framework derives intra-group preference rankings based on Reward Model outputs, while incorporating correctness and length constraints. It then integrates relative rewards using either HRR or PRR, contingent on the availability of rule-based correctness rewards. Finally, the advantage estimation process accounts for correctness consistency and applies a clipping mechanism to handle contradictory samples.}
\label{fig:overview}
\vspace{-10pt}
\end{figure*}

\section{Methodology}

To address the instability of absolute scoring in group-based optimization, we propose \textbf{Reinforcement Learning with Relative Rewards (RLRR)}. This framework anchors advantage estimation in intra-group rankings, providing robust relative signals that align with the comparative nature of the learning objective.

\subsection{Reinforcement Learning with Relative Rewards}

We introduce RLRR, a framework that integrates intra-group relative quality rankings into group-based reinforcement learning, such as GRPO. This approach aims to mitigate gradient vanishing when reward variance collapses within the sampled response set $\{o_i\}_{i=1}^G$. Formally, given the group of responses, we assign each response a rank $r_i \in \{1, \dots, r_{\text{max}}\}$ based on its relative quality, where $r_i=1$ denotes the best response and $r_{\max}$ represents the maximum rank index. We then synthesize these ranks into the final reward via a reward-shaping function $f(\cdot)$ for advantage computation. Depending on the availability of ground truth verification, we devise specific strategies to incorporate these relative rankings as either a fine-grained supplement to rule-based feedback or as the primary reward signal.

\textbf{Hybrid Relative Reward (HRR).}
Tailored for tasks with verifiable outcomes (e.g., mathematical reasoning), HRR preserves the authoritative ground truth signal while introducing fine-grained preference information to resolve tie-breaking scenarios. We define the hybrid reward as a bounded correction to the binary rule-based score $s_i^{\text{rule}}$:
\begin{equation}
    \label{eq:hrr_formula}
    s^{\text{rank}}_i = f_{\text{HRR}}\!\left(s_i^{\text{rule}}, r_i\right) = s_{i}^{\text{rule}} + \tau \cdot \tanh\left(\frac{r_{\max}}{r_i} - 1\right),
\end{equation}
where $\tau$ controls the magnitude of the rank-based adjustment. The hyperbolic tangent function provides a non-linear incentive that significantly boosts top-ranked responses while naturally limiting the correction range. This design ensures that the relative signal effectively differentiates samples sharing identical correctness labels without overriding the primary ground truth objective.

\textbf{Pure Relative Reward (PRR).}
In tasks lacking reliable ground truth, absolute scores from reward models frequently exhibit high variance and range instability. PRR addresses this by replacing raw scalar evaluations with a normalized, rank-centric metric. We define the reward for the $i$-th response with rank $r_i$ as:
\begin{equation}
    \label{eq:prr_formula}
    s^{\text{rank}}_i = f_{\text{PRR}}\!\left(s_i^{\text{rule}}, r_i\right)
     = \frac{r_{\max} - r_i}{r_{\max} - 1}.
\end{equation}
This linear mapping projects the intra-group ranking into a fixed $[0, 1]$ interval. By decoupling the reward signal from the absolute magnitude of model outputs, PRR ensures that advantage estimation depends solely on relative ordering, thereby stabilizing training dynamics against the inherent score shifting of reward models.

\textbf{Correctness-Aware Advantage Clipping.}
While relative advantages enable fine-grained learning, they may occasionally diverge from absolute correctness. A critical misalignment occurs when a valid solution receives a negative advantage simply because it ranks lower within a high-performing group. To prevent the policy from being penalized for generating correct outputs, we introduce a clipping strategy for the raw advantage $A_i$:
\begin{equation}
    \label{eq:adv_clip}
    A_i^{\text{clip}} =
    \begin{cases}
        \max(A_i, \xi^-), & \text{if } o_i \text{ is correct}, \\
        \min(A_i, \xi^+), & \text{if } o_i \text{ is incorrect},
    \end{cases}
\end{equation}
where $\xi^-$ and $\xi^+$ serve as safety margins. This mechanism restricts the magnitude of penalties for valid but suboptimal responses, ensuring that the model learns to differentiate quality nuances without compromising its fundamental capability to generate correct outcomes due to excessive discouragement.

\subsection{Ranking Reward Model}

\begin{figure}[!t]
	\centering
	\includegraphics[width=\columnwidth]{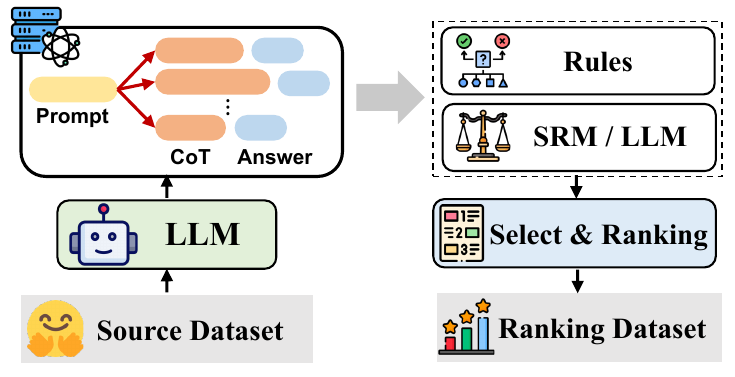}
	\caption{Training-data processing pipeline for the Ranking Reward Model.}
	\label{fig:rrm_data}
     \vspace{-15pt}
\end{figure}

SRMs typically employ a pointwise scoring approach, evaluating responses in isolation to assign absolute scalar values. Lacking comparative context, these independent scores may result in inaccurate rankings when converted to an order. To address this, we introduce the \textbf{Ranking Reward Model (Ranking RM)}. Unlike SRMs, the Ranking RM accepts a list of responses as a collective input and directly outputs their relative ordering. By processing candidates within a shared context, it yields a robust ranking signal that is more reliable than sorting independent scalar scores.

We instantiate the Ranking RM using a pretrained LLM backbone equipped with a classification head to predict ranking permutations, optimized via cross-entropy loss. To construct high-quality training data, we follow the pipeline illustrated in ~\Cref{fig:rrm_data}. Specifically, we establish a hierarchical ranking structure where responses with verified correct outcomes strictly outrank incorrect ones. To determine the relative order of responses with identical correctness labels, we primarily utilize scores from an SRM. However, if the SRM contradicts the ground truth by assigning higher scores to incorrect responses than to correct ones, we discard the unreliable SRM signal and employ a stronger LLM to derive the final rankings.

\subsection{Hierarchical Re-ranking}

Initial rankings derived from reward models may occasionally diverge from ground truth verification. To rectify these potential misalignments, we implement a lexicographical re-ranking strategy to strictly align model predictions with correctness and conciseness constraints. This mechanism primarily prioritizes rule-based correctness to ensure correct solutions outrank incorrect ones, followed by length re-ranking to counteract the verbosity bias inherent in reward models and favor concise responses. We formalize this ranking objective using a coarse-grained discretization function:
\begin{equation}
\label{length_rank}
\mathcal{B}_i =
\begin{cases}
\left\lfloor \dfrac{\ell_i}{\lambda} \right\rfloor, & \text{if } o_i \text{ is correct}, \\
+\infty, & \text{if } o_i \text{ is incorrect},
\end{cases}
\end{equation}
where $\ell_i$ represents the length of response $o_i$ and $\lambda$ serves as a hyperparameter controlling the bin granularity. Responses are sorted in ascending order of $\mathcal{B}_i$. Consequently, among correct responses falling within the same length bin, the final relative order follows the original preference predicted by the reward model. This hierarchical strategy guarantees that the final reward signal aligns with validity and efficiency constraints while preserving the fine-grained quality distinctions captured by the reward model.

{
\begin{algorithm}[H]
  \small
  \caption{RLRR on Verifiable Task}
  \textbf{Input} 
  policy $\pi_{\theta}$, dataset $\mathcal{D}$, rule-based verifier $R_{\phi}$, reward model $R_{\psi}$, group size $G$.
  
\begin{algorithmic}[0]
\FOR{step = 1, \dots, M}
  \STATE Sample a batch $\mathcal{D}_b$ from $\mathcal{D}$ and set $\pi_{\theta_{\text{old}}} \leftarrow \pi_{\theta}$
    \FOR{each $q \in \mathcal{D}_b$}
      \STATE \textbf{Rollout:} Sample a group of responses $\{o_i\}_{i=1}^G \sim \pi_{\theta_{\text{old}}}(q)$.
      \STATE \textbf{Reward Calculation:} Compute correctness scores $\bm s^{\text{rule}}$ and raw relative ranks $\bm r^{\text{raw}}$  using $R_{\phi}$ and $R_{\psi}$.
      \STATE \textbf{Hierarchical Re-ranking:} Derive global ranks $r_i$ by lexicographically sorting the tuple $(\bm s^{\text{rule}}, \mathcal{B}, \bm r^{\text{raw}})$.
    \STATE \textbf{Reward Shaping:}Compute shaping rewards $\bm s^{\text{rank}} $ using the rank-mapping function $ f\!\left(\bm s^{\text{rule}}, \bm r\right)$.
      \STATE \textbf{Advantage Estimation:}  Compute group advantages and apply correctness-aware clipping via \Cref{eq:adv_clip}.
      \STATE \textbf{Policy update:} Update the policy $\pi_{\theta}$ by maximizing the GRPO objective (\Cref{eq:grpo_obj}).
    \ENDFOR
\ENDFOR
\end{algorithmic}
\textbf{Output:} The final policy $\pi_{\theta}$
\label{alg:rank-grpo}
% \vspace{-20pt}
\end{algorithm}
}
\vskip -15pt

Since the Ranking RM evaluates $n$ responses simultaneously, we set the GRPO group size $G$ as a multiple of $n$ and partition the responses into $G/n$ subgroups. As illustrated in Figure~\ref{fig:overview}, we apply the Ranking RM to each subgroup to generate local raw ranks $r_i^{\text{raw}}$ and then perform the hierarchical re-ranking to derive the final global ranks $r_i$ for all $G$ responses. The complete RLRR is formalized in \Cref{alg:rank-grpo}. This integration of intra-group relative ordering effectively mitigates gradient vanishing induced by sparse rewards and circumvents the reliability limitations of absolute scoring. Furthermore, it facilitates priority-aware multi-objective optimization by synthesizing correctness, efficiency, and reasoning quality into a unified ranking signal.

\section{Experiments}

To evaluate the effectiveness of RLRR and the Ranking RM, we conduct experiments across three distinct dimensions:
\begin{itemize}[noitemsep,topsep=3pt]
    \item \textbf{Verifiable Reasoning Tasks}: We assess performance on mathematical and logical reasoning benchmarks where explicit ground truth is available for verification.
    \item \textbf{Open-ended Writing Tasks}: We utilize WritingBench~\citep{wu2025writingbench} to evaluate generation quality across a diverse range of domains and writing styles.
    \item \textbf{Reward Model Evaluation}: We benchmark the discriminative capability of the Ranking RM using reward-guided test-time scaling~\citep{zou2025reasonflux}.
\end{itemize}

\subsection{Verifiable Reasoning Tasks}

\textbf{Baselines.} 
We conduct our experiments on DeepSeek-R1-Distill-Qwen-1.5B and DeepSeek-R1-Distill-LlaMA-8B~\citep{guo2025deepseek}. Our primary comparison is against four recent state-of-the-art reinforcement learning methods: 
(1) GRPO~\citep{shao2024deepseekmath}, 
(2) Dr.GRPO~\citep{liu2025understanding}, 
(3) DAPO~\citep{yu2025dapo},
and (4) GPG~\citep{chu2025gpg}.
The Ranking RM is trained on 25k data using Qwen2.5-7B-Instruct-1M~\citep{qwen2.5-1m}.

\textbf{Datasets.}
For RL training, we use approximately 16k mathematics and logic samples filtered by difficulty from the GURU dataset~\citep{cheng2025revisiting}, and for the 8B model, we additionally include the Open-RS dataset~\citep{dang2025reinforcementlearningreasoningsmall}. For ablation and analysis experiments, we train the 1.5B model using the SimpleRL dataset~\citep{zeng2025simplerl}. For evaluation, we employ several challenging mathematical and logical reasoning benchmarks to assess our models' performance. Detailed descriptions and references for all evaluation datasets are provided in Appendix~\ref{app:datas}.

{

\newcommand{\res}[2]{#1{\scriptsize\color{gray}$\pm$#2}} 
\newcommand{\bres}[2]{\textbf{#1}{\scriptsize\color{gray}$\pm$#2}} % 加粗 (Best)
\newcommand{\ures}[2]{\underline{#1}{\scriptsize\color{gray}$\pm$#2}} % 下划线 (Second best)
\newcommand{\len}[2]{#1{\scriptsize\color{gray}$\pm$#2}} % 长度列占位
\newcommand{\blen}[2]{\textbf{#1}{\scriptsize\color{gray}$\pm$#2}} % 长度列占位
\newcommand{\ulen}[2]{\underline{#1}{\scriptsize\color{gray}$\pm$#2}} % 长度列占位

\renewcommand{\arraystretch}{1.25} %稍微增加行高，使带上下标的数值不拥挤
\setlength{\tabcolsep}{2.5pt}    %稍微减小列间距，防止表格因为增加了内容而过宽

\begin{table*}[!t]
\centering
\caption{Overall performance on eight competition-level mathematical reasoning benchmarks and two logic reasoning benchmarks. 
\textbf{Bold} and \underline{underlined} indicate the best and second-best performance, respectively.}
\label{tab:math-main}

\resizebox{\textwidth}{!}{%
\begin{tabular}{cl ccccccccc ccc c}
\toprule
% --- 表头第1行 ---
\multicolumn{2}{c}{\multirow{3}{*}{\normalsize \textbf{\textsc{Method}}}} & 
\multicolumn{9}{c}{\textit{\textbf{\textsc{Mathematical Reasoning}}}} & 
\multicolumn{3}{c}{\textbf{\textit{\textsc{Logic Reasoning}}}} & 
\multirow{3}{*}{\shortstack{\textbf{Avg} \\ \textbf{Len.}}} \\
\cmidrule(lr){3-11} \cmidrule(lr){12-14}

% --- 表头第2行 ---
\multicolumn{2}{c}{} & 
\textbf{AIME 24} & \textbf{AIME 25} & \textbf{MATH 500} & \textbf{GSM8K} & \textbf{Olympiad} & \textbf{GaoKao} & \textbf{Minerva} & \textbf{AMC} & 
\multicolumn{1}{c}{\multirow{2}{*}{\textbf{Avg}}} & 
\textbf{Zebra} & \textbf{Ordering} & 
\multicolumn{1}{c}{\multirow{2}{*}{\textbf{Avg}}} & \\
\cmidrule(lr){3-3} \cmidrule(lr){4-4} \cmidrule(lr){5-5} \cmidrule(lr){6-6} \cmidrule(lr){7-7} \cmidrule(lr){8-8} \cmidrule(lr){9-9} \cmidrule(lr){10-10} \cmidrule(lr){12-12} \cmidrule(lr){13-13}

% --- 表头第3行 (Metric) ---
\multicolumn{2}{c}{} & 
\textit{Avg@32} & \textit{Avg@32} & \textit{Avg@4} & \textit{Avg@4} & \textit{Avg@4} & \textit{Avg@4} & \textit{Avg@4} & \textit{Avg@16} & & 
\textit{Avg@4} & \textit{Avg@4} & & \textit{Tokens} \\
\midrule

\multirow{7}{*}{\rotatebox{90}{\textbf{\textit{DeepSeek-Qwen-1.5B}}}}

& Baseline & 26.4 & 21.8 & 83.2 & 86.1 & 41.7 & 71.3 & 26.3 & 61.4 & 52.3 & 0.7 & 14.0 & 7.4 & 10050 \\
% &  &  &  &  &  &  &  &  &  &  &  &  &  \\
 & GRPO & 29.3 & 23.4 & 82.7 & 85.7 & 42.8 & 72.7 & 27.6 & 63.0 & 53.4 & 2.5 & 22.4 & 12.5 & 6943 \\
 & Dr.GRPO & 29.1 & 23.8 & 83.0 & 85.6 & 43.7 & 73.6 & 27.0 & 63.6 & 53.7 & 3.0 & 20.4 & 11.7 & 7025 \\
 & DAPO & 28.1 & 22.9 & 83.5 & 86.3 & 44.1 & 73.7 & 28.0 & 65.3 & 54.0 & {\ul 6.8} & 28.4 & 17.6 & 9536 \\
 & GPG & 30.4 & \underline{24.2} & {\ul 84.2} & {\ul 86.4} & 44.2 & 73.4 & \textbf{28.3} & 62.8 & 54.2 & 3.8 & 22.2 & 13.0 & 8548 \\
\cmidrule(lr){2-15}
% & RLRR(H)
% &  &  &  &  &  &  &  &  &  &  &  &  &  \\
% & RLRR(P)
% &  &  &  &  &  &  &  &  &  &  &  &  &  \\
 & RLRR(H) & \underline{31.1} & \textbf{24.5} & \textbf{84.6} & \textbf{86.5} & {\ul 47.0} & \textbf{74.6} & 28.0 & {\ul 66.8} & {\ul 55.4} & \textbf{10.1} & \textbf{39.5} & \textbf{24.8} & \textbf{6484} \\
 & RLRR(P) & \textbf{32.5} & 23.2 & 83.8 & 86.2 & \textbf{47.8} & {\ul 74.4} & {\ul 28.1} & \textbf{68.3} & \textbf{55.5} & 6.1 & {\ul 31.4} & {\ul 18.8} & {\ul 6727} \\

\midrule
% \midrule

% ================= R1-Distill-LLaMA-8B =================
\multirow{7}{*}{\rotatebox{90}{\textbf{\textit{DeepSeek-LLaMA-8B}}}}
& Baseline     
& 46.8 & 29.3 & 88.4 & 91.0 & 51.6 & 79.3 & 28.9 & 77.3 & 61.6 & 9.0 & 55.8 & 32.4 & 7871 \\
 % &  &  &  &  &  &  &  &  &  &  &  &  &  \\
& GRPO & 48.1 & 30.0 & 88.9 & 91.3 & 53.1 & 80.4 & 31.3 & 80.2 & 62.9 & 30.8 & 77.7 & 54.3 & 6567 \\
& Dr.GRPO & {\ul 48.8} & 29.8 & {\ul 89.4} & 91.5 & 53.5 & 79.0 & 30.8 & {\ul 80.9} & 63.0 & 23.5 & 74.5 & 49.0 & 6466 \\
& DAPO & 46.8 & {\ul 34.2} & 89.3 & \textbf{92.1} & {\ul 58.4} & {\ul 81.9} & 32.0 & 79.0 & {\ul 64.2} & {\ul 37.3} & \textbf{89.3} & \underline{63.3} & 5871 \\
& GPG & 47.6 & 30.1 & 89.0 & 91.3 & 54.1 & 80.4 & 31.3 & 80.2 & 63.0 & 24.6 & 78.8 & 51.7 & 6920 \\
\cmidrule(lr){2-15}
% & RLRR(H)
% &  &  &  &  &  &  &  &  &  &  &  &  &  \\
% & RLRR(P)
% &  &  &  &  &  &  &  &  &  &  &  &  &  \\
& RLRR(H) & \textbf{50.5} & \textbf{35.4} & \textbf{91.4} & \textbf{92.1} & \textbf{59.0} & \textbf{82.1} & \textbf{33.5} & \textbf{83.0} & \textbf{65.9} & \textbf{40.4} & \underline{89.2} & \textbf{64.8} & \textbf{5025} \\
& RLRR(P) & {\ul 48.8} & 34.1 & {\ul 89.4} & {\ul 91.7} & 54.8 & 80.5 & {\ul 32.4} & {\ul 80.9} & 64.1 & 35.8 & 88.0 & 61.9 & {\ul 5287} \\
\bottomrule
\end{tabular}
}
\end{table*}
}

\textbf{Performance.} 
As presented in \Cref{tab:math-main}, RLRR demonstrates robust performance on both mathematical and logical reasoning benchmarks, with both the HRR and PRR variants yielding effective results. Notably, beyond achieving high accuracy, RLRR maintains superior inference efficiency, requiring the lowest token consumption among all baseline methods. This efficiency underscores the efficacy of ranking-based optimization in balancing performance with computational cost. Specifically, HRR, which anchors advantage estimation to ground truth rules, yields the most substantial and stable improvements on verifiable tasks. In contrast, PRR achieves slightly lower performance, a result attributed to its exclusion of direct rule-based supervision. Collectively, these findings validate the effectiveness of RLRR in enhancing both mathematical and logical reasoning capabilities across diverse domains.

\subsection{Open-ended Writing Tasks}

\renewcommand{\arraystretch}{1.2}{
\begin{table*}[!t]
    \centering
    \caption{Overall performance on Writing Bench, comparing reward scores and reward ranks. 
The `$\hookrightarrow$` symbol denotes a variant of the method listed directly above. 
\textbf{Bold} and \underline{underline} mark the best and second-best results, respectively.}
    \label{tab:writing_bench}
    \small
    \resizebox{\textwidth}{!}{%
    \begin{tabular}{l *{7}{S[table-format=2.2]}}
        \toprule
        \multirow{2}{*}{\textsc{\textbf{Method}}} & {\multirow{2}{*}{\textsc{\textbf{Overall}}}} & {\textsc{\textbf{Academic}}} & {\textsc{\textbf{Finance}}} & {\textsc{\textbf{Politics}}} & {\textsc{\textbf{Literature}}} & {\multirow{2}{*}{\textsc{\textbf{Education}}}} & {\textsc{\textbf{Advertising}}} \\
        & & {\textsc{\textbf{\& Engineering}}} & {\textsc{\textbf{\& Business}}} & {\textsc{\textbf{\& Law}}} & {\textsc{\textbf{\& Arts}}} & & {\textsc{\textbf{\& Marketing}}} \\
        \midrule
        
        % --- Baselines ---
        Qwen3-1.7B & 70.06 & 72.60 & 71.17 & 70.99 & 63.22 & 73.52 & 70.27 \\
        SFT & 70.90 & 73.17 & 70.89 & 71.47 & 65.75 & 74.68 & 71.09 \\
        % \midrule
        \cmidrule(lr){1-8}

        % --- Reward Models Comparison ---
        % Skywork-8B and its Ranking variant
        Skywork-8B & 72.88 & 74.56 & 72.81 & 72.40 & 69.68 & 76.00 & 73.42 \\
        $\hookrightarrow$ \textit{RLRR} & \chg{73.64}{0.76} & \chg{75.25}{0.69} & \chg{73.71}{0.90} & \chg{73.77}{1.37} & \chg{69.81}{0.13} & \chgu{77.22}{1.22} & \chg{73.57}{0.15} \\
        \cmidrule(lr){1-8} 

        % URM-8B and its Ranking variant
        URM-8B & 73.12 & 75.14 & 73.65 & 73.47 & 69.06 & 76.16 & 72.25 \\
        $\hookrightarrow$\textit{RLRR} & \chgu{74.71}{1.59} & \chgu{76.46}{1.32} & \chgu{75.47}{1.82}  & \chgu{75.30}{1.83} & \chgu{70.86}{1.80} & \chg{77.08}{0.92} & \chgu{73.72}{1.47} \\
        \cmidrule(lr){1-8}
        % \rowcolor{gray!20}
        Ranking RM(ours) & \textbf{81.33} & \textbf{83.27} & \textbf{82.92} & \textbf{81.68} & \textbf{75.90} & \textbf{84.16} & \textbf{80.96} \\
        \bottomrule
    \end{tabular}
    }
    \vspace{-12pt}
\end{table*}
% \vspace{-15pt}
}

\textbf{Datasets and Models.}
Following the pipeline in \Cref{fig:rrm_data}, we constructed the preference dataset by generating ranked responses for 10k Dolphin-R1\footnote{\url{https://huggingface.co/datasets/QuixiAI/dolphin-r1}} samples using models of varying scales. We then fine-tuned the Qwen3-1.7B backbone~\citep{yang2025qwen3} on a separate set of 22k samples.

\textbf{Baselines.}
We adopt GRPO~\citep{shao2024deepseekmath} as the algorithmic baseline for comparison. In addition, we benchmark the Ranking RM against two state-of-the-art reward models: (1) Skywork-Reward-V2-Llama-3.1-8B~\citep{liu2025skywork}, (2) URM-LLaMa-3.1-8B~\citep{lou2024uncertainty}.

\textbf{Performance.} 
\Cref{tab:writing_bench} presents the performance of RLRR on open-ended tasks. By converting absolute scalar scores from two distinct reward models into relative preferences through the PRR approach, we achieve improvements across most domains. This highlights the effectiveness of relative ranking in enhancing both the stability and performance of GRPO. Moreover, when RLRR is paired with the fine-tuned Ranking RM, it achieves the best results in the majority of domains, underscoring that evaluating the preference order of multiple responses provides a more robust learning signal than scoring individual responses. These findings reinforce the value of ranking-based approaches for improving performance in open-ended tasks.

\subsection{Evaluation of Ranking Reward Model}

\begin{figure*}[!t]
  \centering
  \begin{subfigure}[t]{0.2\textwidth}
    \centering
    \includegraphics[width=\linewidth]{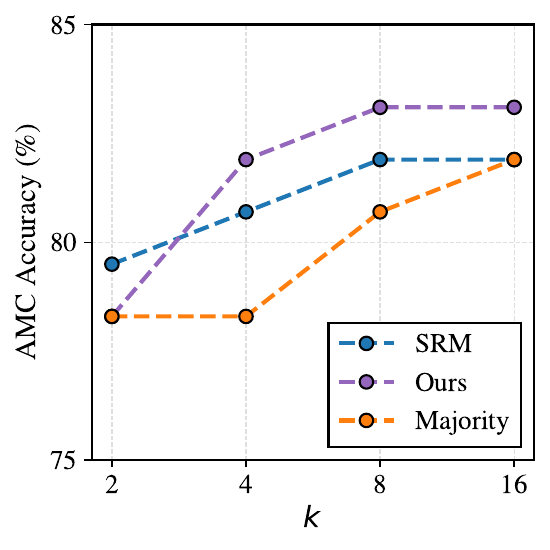}
    \caption[R]{\small Math}
    \label{fig:math_line}
  \end{subfigure}
  \begin{subfigure}[t]{0.2\textwidth}
    \centering
    \includegraphics[width=\linewidth]{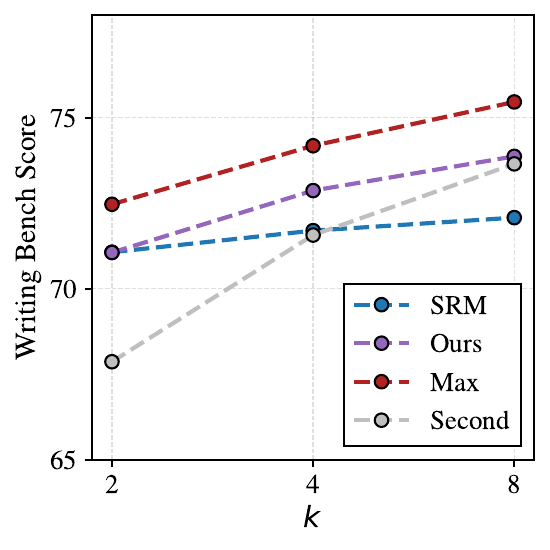}
    \caption[R]{\small Writing}
    \label{fig:writing_line}
  \end{subfigure}
  \begin{subfigure}[t]{0.58\textwidth}
    \centering
    \includegraphics[width=\linewidth]{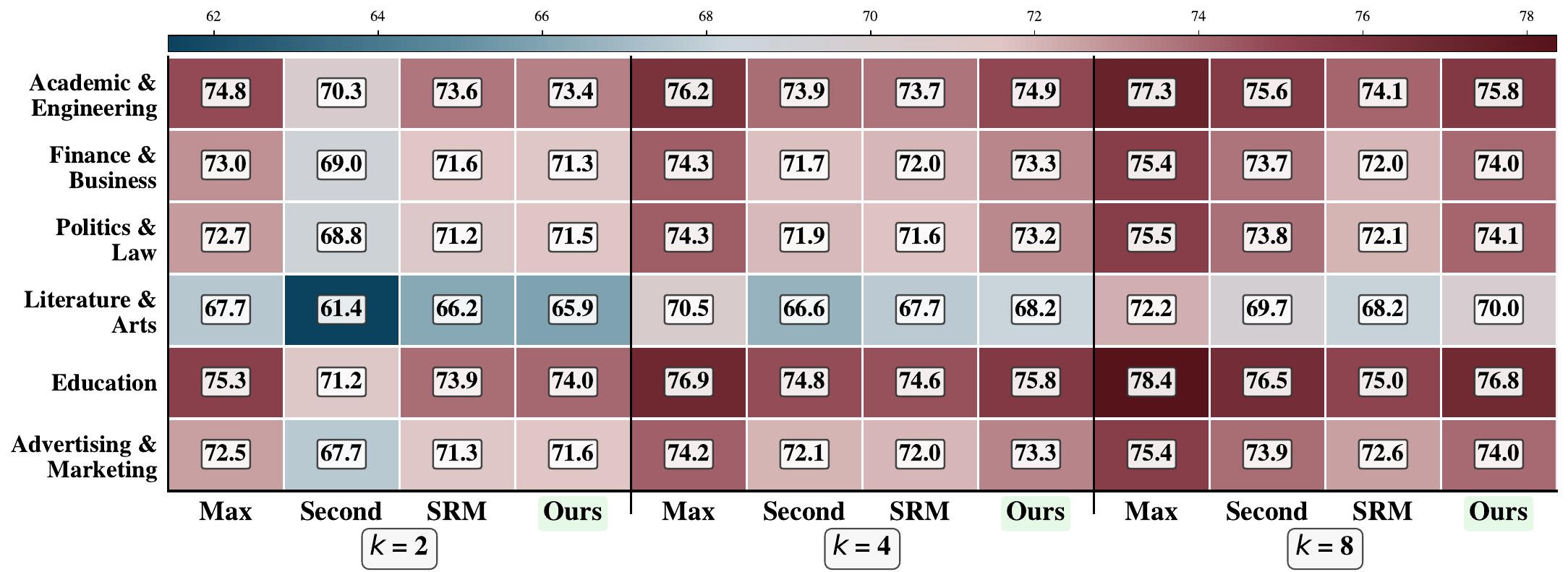}
    \caption[R]{\small Cross-Domain RM Comparison in Writing Benchmark}
    \label{fig:map_writing}
  \end{subfigure}
  \caption{Reward-Guided Best-of-N Test-Time Scaling for Enhanced Inference Performance.}
  \label{fig:reward_eval}
  \vspace{-10pt}
\end{figure*}

\Cref{fig:reward_eval} compares the SRM Skywork-Reward-V2-Llama-3.1-8B~\citep{liu2025skywork} and the Ranking RM under two experimental configurations. Figure 1 demonstrates their performance on mathematical reasoning tasks using the DeepSeek-LlaMA-8B ~\citep{guo2025deepseek}, where we sample \(k\) responses per prompt and select via either SRM or Ranking RM with Majority voting. Results show Ranking RM achieves comparable or superior accuracy to SRM, with the performance gap widening as \(k\) increases.
\Cref{fig:math_line} and \ref{fig:writing_line} illustrate Ranking RM's superior performance in writing tasks evaluated on Qwen3-1.7B~\citep{yang2025qwen3}, where we sample eight responses per instance. The \textit{Second} designation denotes the second-highest-scoring sample. Ranking RM consistently selects higher-quality responses, explaining its significant improvement in compositional tasks.
We attribute these improvements to Ranking RM's focus on relative intra-group quality assessment rather than absolute scoring. Notably, Ranking RM demonstrates strong generalization across domains despite limited training data.

\subsection{Method Analysis}
\label{sec:data_level_main}

We analyze RLRR from multiple perspectives, with detailed results provided in Appendix~\ref{app:full_result}.

\begin{figure*}[!t]
    \centering
    \captionsetup[sub]{skip=0pt} % 子图标题紧凑

    % 第一行三张
    \begin{subfigure}{0.32\textwidth}
        \centering
        \includegraphics[width=\linewidth]{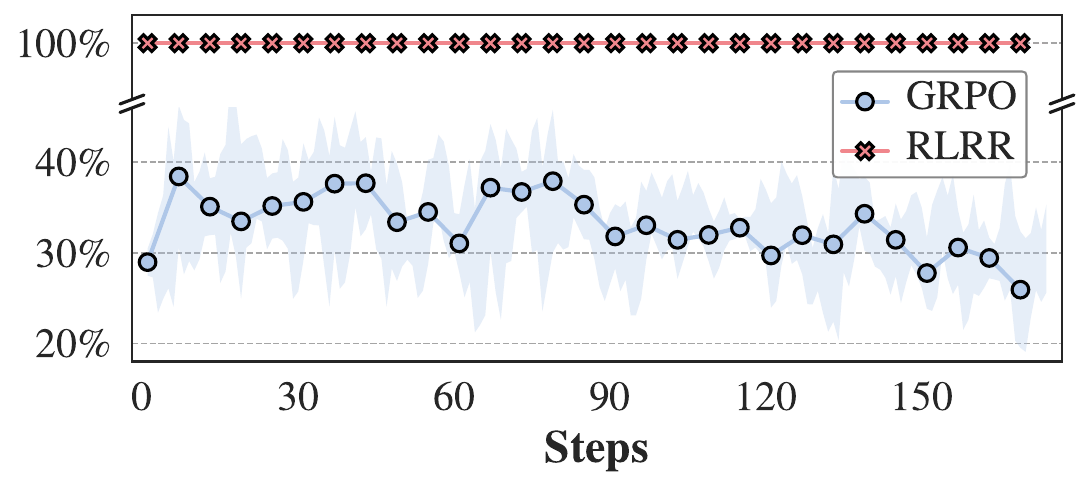}
        \subcaption{\scriptsize Ratio of Effective Easy Prompts}
    \end{subfigure}\hspace{0.01\textwidth}
    \begin{subfigure}{0.32\textwidth}
        \centering
        \includegraphics[width=\linewidth]{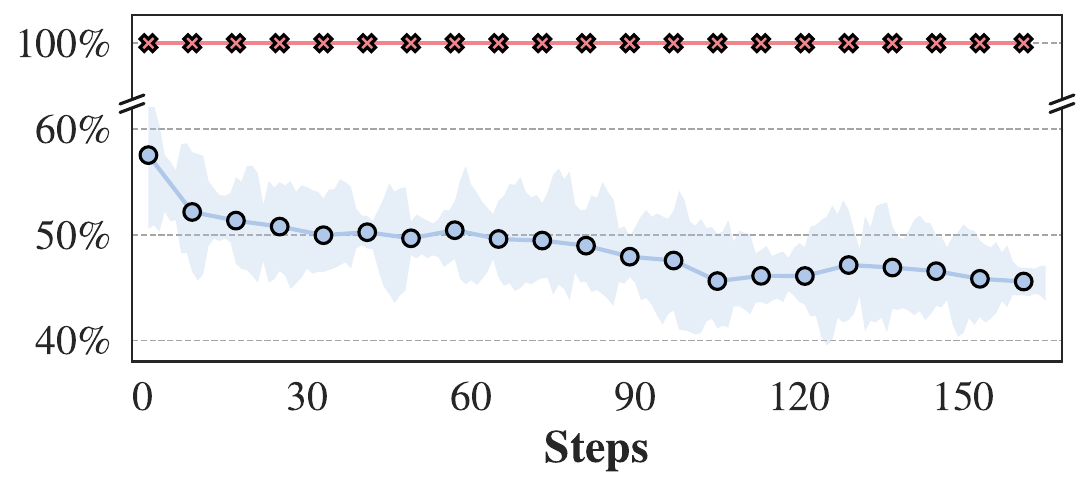}
        \subcaption{\scriptsize Ratio of Effective Medium Prompts}
    \end{subfigure}\hspace{0.01\textwidth}
    \begin{subfigure}{0.32\textwidth}
        \centering
        \includegraphics[width=\linewidth]{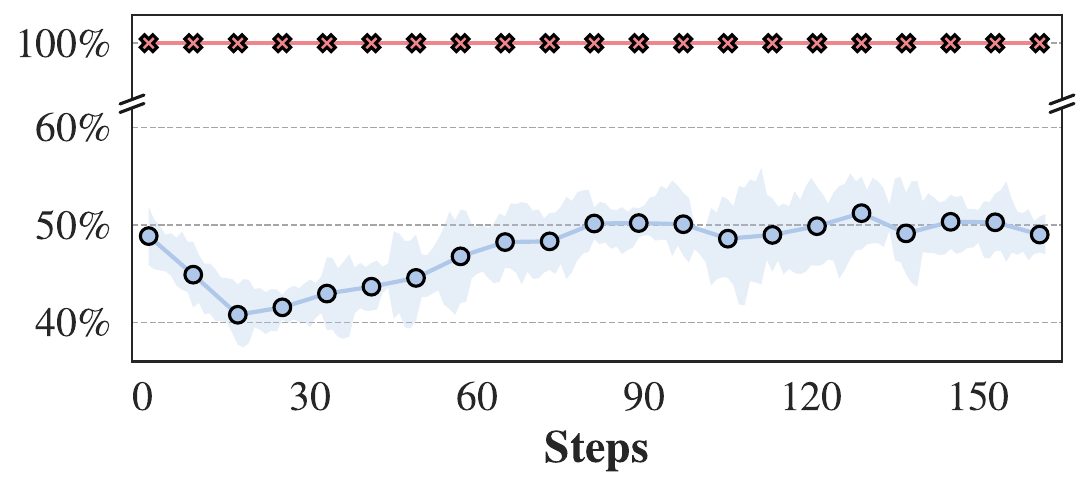}
        \subcaption{\scriptsize Ratio of Effective Hard Prompts}
    \end{subfigure}
    
    \vspace{-0.1em}

    % 第二行三张
    \begin{subfigure}{0.32\textwidth}
        \centering
        \includegraphics[width=\linewidth]{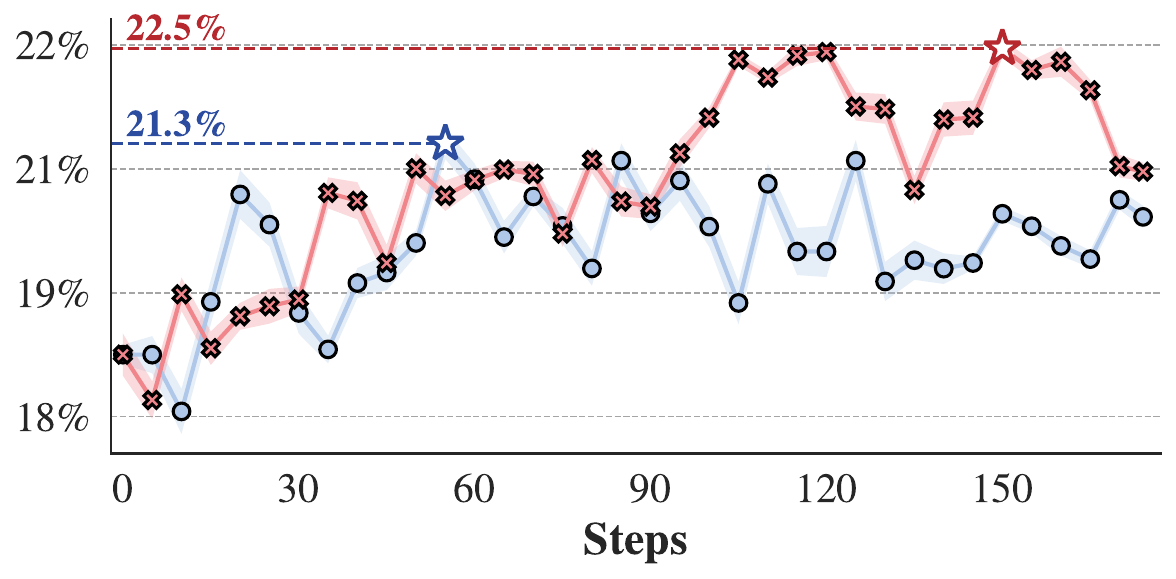}
        \subcaption{\scriptsize AIME2025 Avg@8 of Easy Data}
    \end{subfigure}\hspace{0.01\textwidth}
    \begin{subfigure}{0.32\textwidth}
        \centering
        \includegraphics[width=\linewidth]{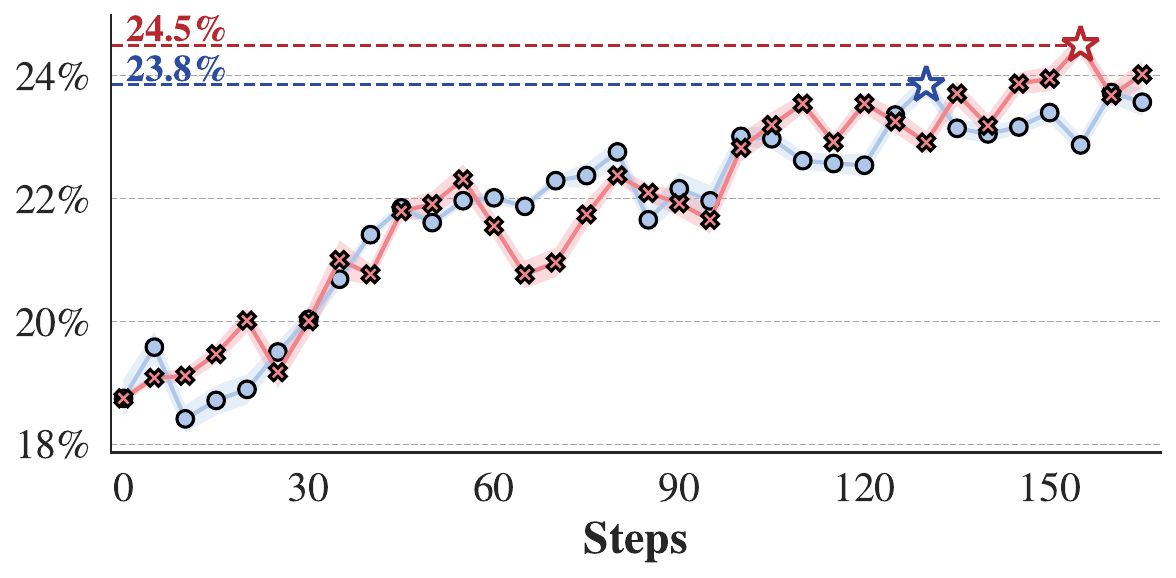}
        \subcaption{\scriptsize AIME2025 Avg@8 of Medium Data}
    \end{subfigure}\hspace{0.01\textwidth}
    \begin{subfigure}{0.32\textwidth}
        \centering
        \includegraphics[width=\linewidth]{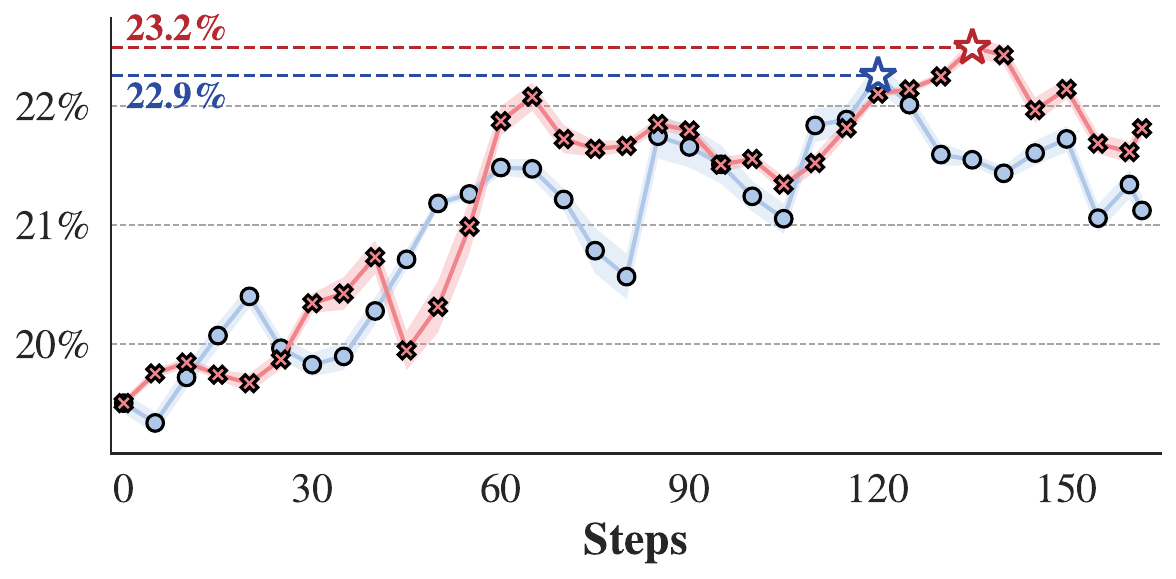}
        \subcaption{\scriptsize AIME2025 Avg@8 of Hard Data}
    \end{subfigure}

    \caption{Comparison of GRPO and RLRR Across Different Data Difficulty Levels.}
    \label{fig:data_level}
    \vspace{-12pt}
\end{figure*}

\textbf{Impact of Dataset Difficulty.}
The dataset difficulty directly affects the proportion of the effective prompts. To examine this, we fine-tune a 1.5B model on three levels: (1) Easy: GSM8k~\citep{cobbe2021gsm8k}, (2) Medium: SimpleRL~\citep{zeng2025simplerl}, and (3) Hard: Open-RS~\citep{dang2025reinforcementlearningreasoningsmall}.
\Cref{fig:data_level} presents the performance of GRPO and RLRR across datasets of varying difficulty. 
We observe that the fraction of effective prompts under GRPO remains relatively low and  exhibits difficulty dependent dynamics, while RLRR maintains full utilization of group-wise information regardless of task complexity.
On the easy dataset, RLRR further improves model performance, whereas GRPO shows limited gains. On medium and hard datasets, both methods benefit from more informative prompts, and RLRR consistently outperforms GRPO. Further analysis of these effects is provided in Appendix~\ref{app:data_level_discuss}.

\begin{table}[!t]
    \centering
    \caption{Ablation study results on Hierarchical Re-ranking. \textbf{Bold} marks the best result in each column.
    \textit{Cor.} means Correctness, and \textit{Len.} means Length.
    % \textcolor{red}{table need to update}
    }
    \vspace{-2pt}
    \label{tab:ablation_rrr}
    \small 
\resizebox{\columnwidth}{!}{%
    \begin{tabular}{lcccccccc} 
        \toprule
        \textbf{\textsc{Method}} & {AIME24} & {AIME25} & {Meth} & {Olym.} & {Mine.} & {AMC} & Avg & Len. \\
        \midrule
        RLRR  & 30.8 & 24.5 & 83.8 & 46.5 & 27.6 & \textbf{69.1} & {47.1} & \textcolor{black}{6013} \\
        w/o \textit{Cor.}     & 28.4 & 23.1 & 83.9 & 44.8 & 27.7 & 66.3 & 45.7 & \textcolor{black}{\textbf{5720}} \\
        w/o \textit{Len.}     & \textbf{31.3} & \textbf{24.7} & \textbf{85.0} & \textbf{47.2} & \textbf{27.9} & 66.8 & \textbf{47.2} & \textcolor{black}{6317} \\
        \bottomrule
    \end{tabular}
}
\vspace{-13pt}
\end{table}

\textbf{Ablation of Hierarchical Re-ranking.}
We analyze the contribution of different components in the re-ranking stage. 
Table~\ref{tab:ablation_rrr} shows that removing correctness consistently degrades performance, confirming its necessity for reliable reasoning. 
Interestingly, relaxing the length constraint slightly improves results on several datasets, suggesting that over-restricting output length may hinder problem-solving flexibility. 
Overall, correctness serves as the primary factor for stability, while length control requires a careful balance between conciseness and expressiveness. 

\begin{figure*}[!t]
    \centering
    %----------------------------------------------------------------------%
    %  图1：Rule-base Rewards
    %----------------------------------------------------------------------%
    \begin{minipage}[t]{0.275\textwidth}
        \centering
        \includegraphics[width=\textwidth]{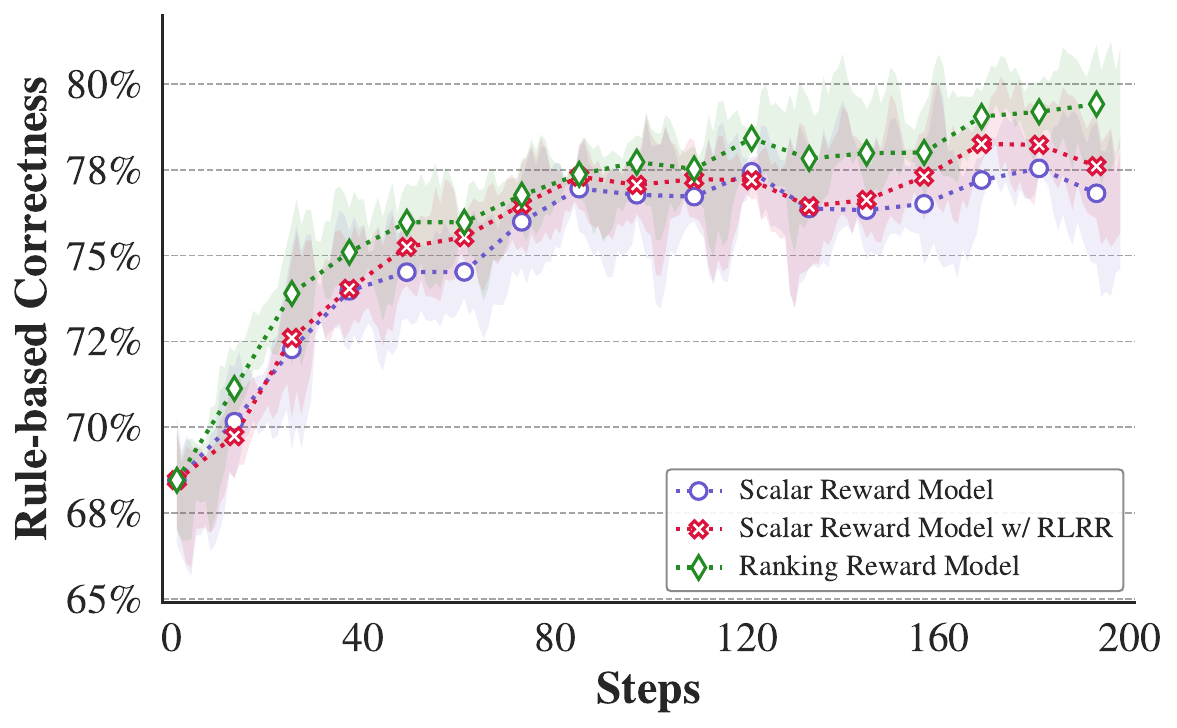}
        \subcaption{\small Rule-base Correctness}\label{fig:rm_rule_line}
    \end{minipage}%
    \hfill % 自动填充间距
    %----------------------------------------------------------------------%
    %  图2：AIME Accuracy
    %----------------------------------------------------------------------%
    \begin{minipage}[t]{0.275\textwidth}
        \centering
        \includegraphics[width=\textwidth]{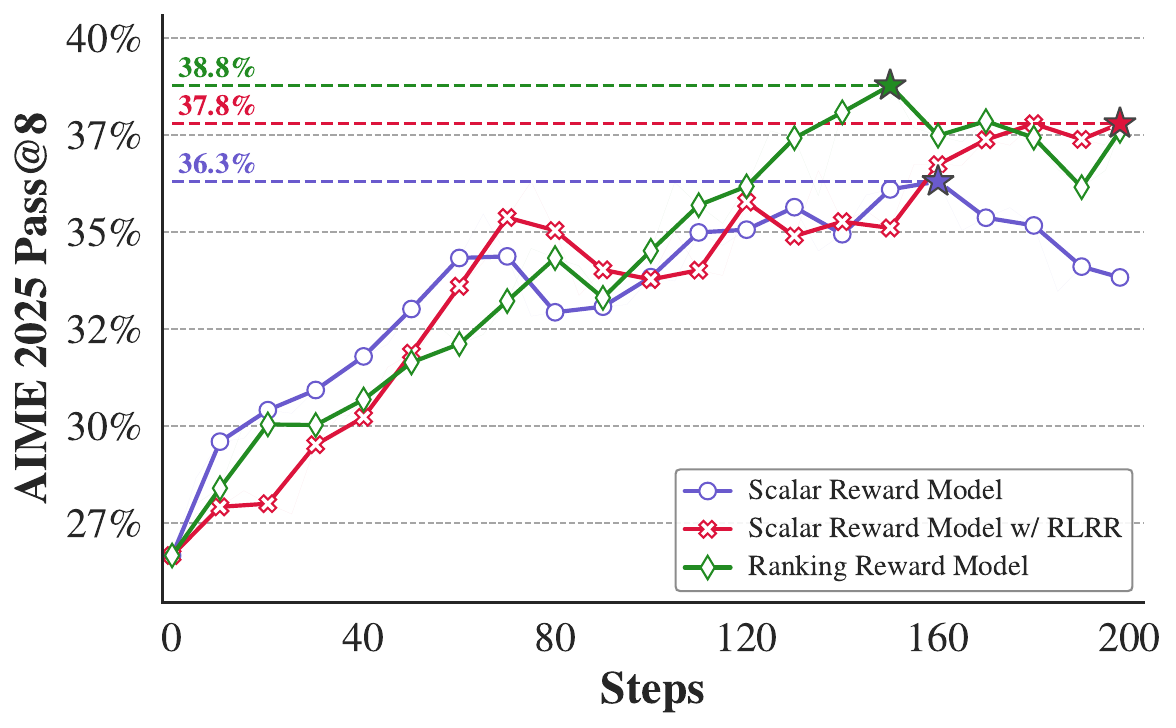}
        \subcaption{\small AIME2025 Accuracy}\label{fig:rm_aime_line}
    \end{minipage}%
    \hfill % 自动填充间距
    %----------------------------------------------------------------------%
    %  图3：Performance Bar
    %----------------------------------------------------------------------%
    \begin{minipage}[t]{0.44\textwidth}
        \centering
        \includegraphics[width=\textwidth]{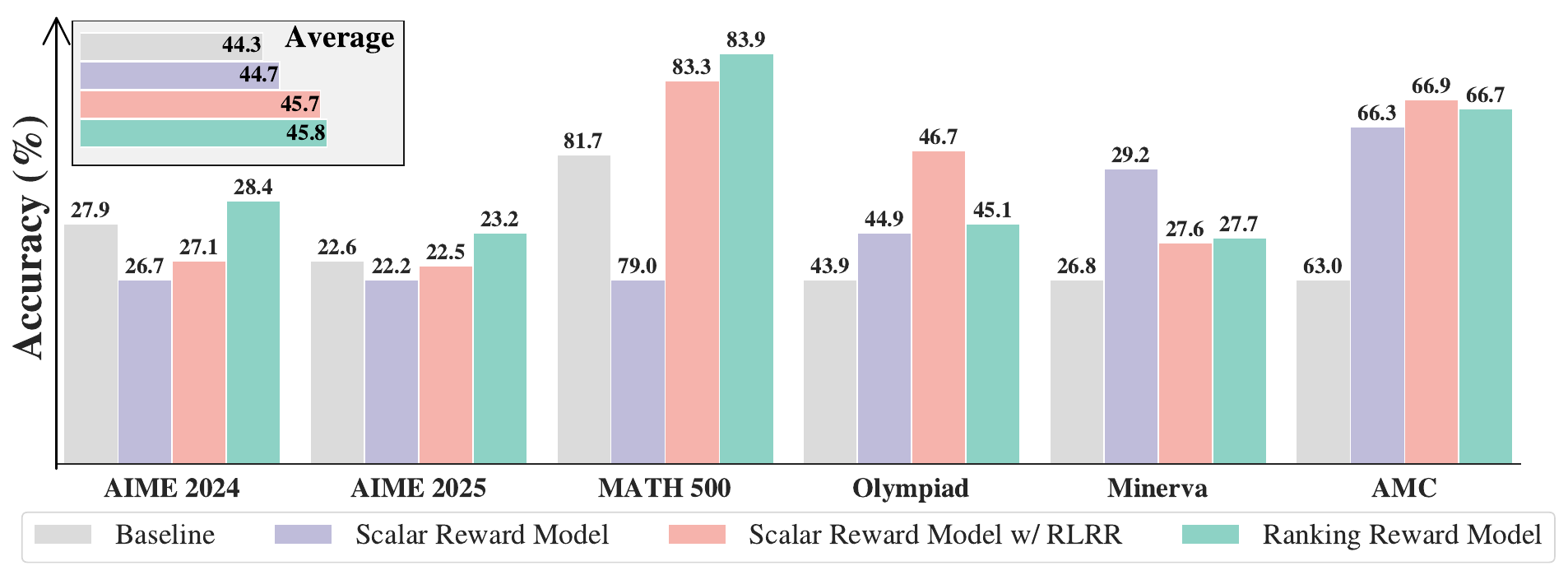}
        \subcaption{\small Performance with Different Reward Models}\label{fig:rm_bar}
    \end{minipage}
    
    \caption{Training dynamics and reward analyses of RLRR excluding correctness rewards.}
    \label{fig:deeper_eval}
\end{figure*}

\begin{table}[!t]
    \centering
    \caption{Performance comparison of RLRR across different group-based reinforcement learning algorithms. }
    \label{tab:rank_more_algo}
    \small 
    \resizebox{\columnwidth}{!}{%
        \begin{tabular}{
            l
            *{6}{S[
                table-format=2.1, 
                table-auto-round,
                detect-weight=true, 
                mode=match,
                table-space-text-post=\hspace{2em}
            ]}
        } 
        \toprule
        \textbf{\textsc{Method}} & {\textbf{AIME24}} & {\textbf{AIME25}} & {\textbf{Minerva}} & {\textbf{Olympiad}} & {\textbf{AMC\quad}} & {\textbf{Avg\qquad}} \\
        \midrule
        CISPO  & 27.7 & 22.8 & 26.6 & 42.3 & 61.3 & 36.1 \\
        \rowcolor{gray!20}
        \, w/ RLRR         
        & \chg{29.8}{2.1} 
        & \chg{23.4}{0.6} 
        & \chg{27.5}{0.9} 
        & \chg{42.2}{-0.1} 
        & \chg{62.7}{1.4} 
        & \chg{37.1}{1.0}  \\
        RLOO  & 28.1 & 22.4 & 26.3 & 42.1 & 61.4 & 36.1 \\
        \rowcolor{gray!20}
        \, w/ RLRR     
        & \chg{28.9}{0.8} 
        & \chg{23.2}{0.8} 
        & \chg{27.4}{1.1} 
        & \chg{43.2}{1.1} 
        & \chg{62.9}{1.5} 
        & \chg{37.1}{1.0}  \\
        \bottomrule
    \end{tabular}
}
\vspace{-12pt}
\end{table}

\begin{table*}[!t]
    \centering
    \caption{Overall performance on eight competition-level mathematical reasoning benchmarks of pre-trained model. \textbf{Bold} and \underline{underlined} indicate the best and second-best performance, respectively.
    }
    \label{tab:base_model_rl}
    \small 
\resizebox{\textwidth}{!}{%
    \begin{tabular}{lccccccccc c} 
        \toprule
        \textbf{\textsc{Method}} & \textbf{\textsc{Reward}} & {\textbf{AIME24}} & {\textbf{AIME25}} & {\textbf{MATH500}}  & {\textbf{Olympiad}} & {\textbf{GaoKao}} & {\textbf{Minerva}} & {\textbf{AMC}} & \textbf{Avg} & \textbf{Avg Len.} \\
        \midrule
        Qwen-2.5-7B & - & 3.6 & 1.1 & 40.2 & 16.2 & 33.2 & 14.9 & 22.4 & 22.8 & 915 \\
        GRPO  & Absolute & 8.2 & \underline{3.8} & 63.2 & 27.3 & 55.3 & 24.9 & 36.1 & 31.3 & 857 \\
        % DAPO  & Rule  &  &  &  &  &  &  &  &  &  \\
        DAPO  & Absolute  & 7.6 & 3.6 & \underline{63.4} & 27.3 & 55.5 & 24.2 & 37.0 & 31.2 & 874 \\
        \cmidrule(lr){1-11}
        % \rowcolor{gray!20}
        RLRR w/ SRM  & Relative &  \underline{8.6} & \textbf{3.9} & \textbf{63.9} & \underline{27.4} & \textbf{56.0} & \textbf{25.8} & \underline{37.5} & \underline{31.9} & \textbf{827} \\
 % &  &  &  &  &  &  &  &  &  \\
        % \rowcolor{gray!20}
        RLRR  & Relative  & \textbf{10.0} & 3.6 & 63.1 & \textbf{27.9} & \underline{55.8} & \underline{25.5} & \textbf{38.8} & \textbf{32.1} & \underline{852} \\
        \bottomrule
    \end{tabular}
}
\vspace{-10pt}
\end{table*}

\textbf{Comparison of Reward Shaping Strategies.}
We further investigate strategies for using reward models. As shown in \Cref{fig:rm_bar}, SRMs suffer from instability, occasionally leading to performance degradation on specific benchmarks. In contrast, converting absolute scores into relative rankings significantly stabilizes the reward signal and yields consistent performance gains. \Cref{fig:rm_rule_line} and \Cref{fig:rm_aime_line} illustrate this effect from the perspective of training dynamics, highlighting how relative rewards mitigate signal fluctuations. Furthermore, adopting our proposed Ranking RM delivers the most substantial improvements, providing empirical evidence that listwise ranking signals offer superior reliability compared to pointwise scalar values. These findings confirm that relative ranking-based rewards effectively enhance both training stability and final model performance.

\textbf{Generalizability to Other Group-based Algorithms.}
To validate the versatility of our framework, we investigate the impact of relative rewards on two additional group-based reinforcement learning algorithms: CISPO~\citep{chen2025minimax} and RLOO~\citep{ahmadian2024rloo}. As detailed in \Cref{tab:rank_more_algo}, integrating relative rewards yields consistent performance improvements for both baselines. These results confirm that the benefits of the RLRR paradigm extend beyond GRPO, demonstrating its robust effectiveness across different group-based optimization methodologies.

\begin{figure}[!t]
  \centering
    \begin{minipage}{\columnwidth} 
        \centering
        \begin{minipage}{0.48\columnwidth}
            \includegraphics[width=\columnwidth]{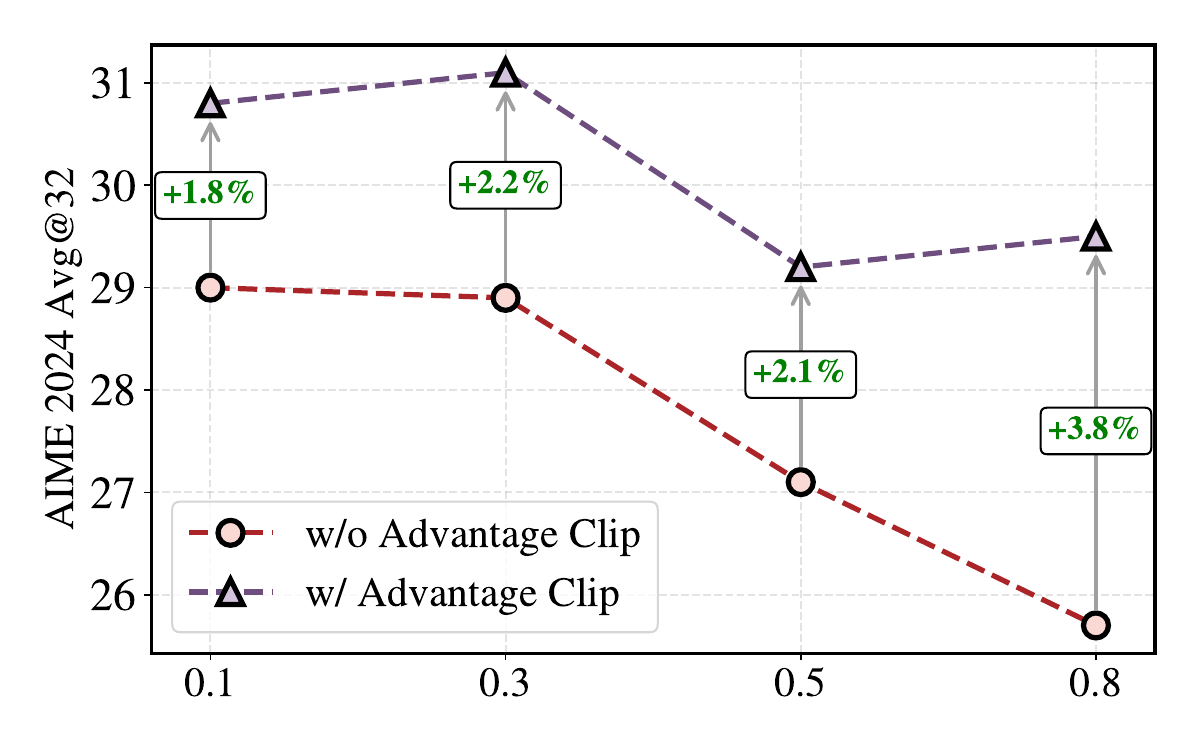}
            \subcaption{\small Analysis of $\tau$}
            \label{fig:tau}
        \end{minipage}
        \begin{minipage}{0.48\columnwidth}
            \includegraphics[width=\columnwidth]{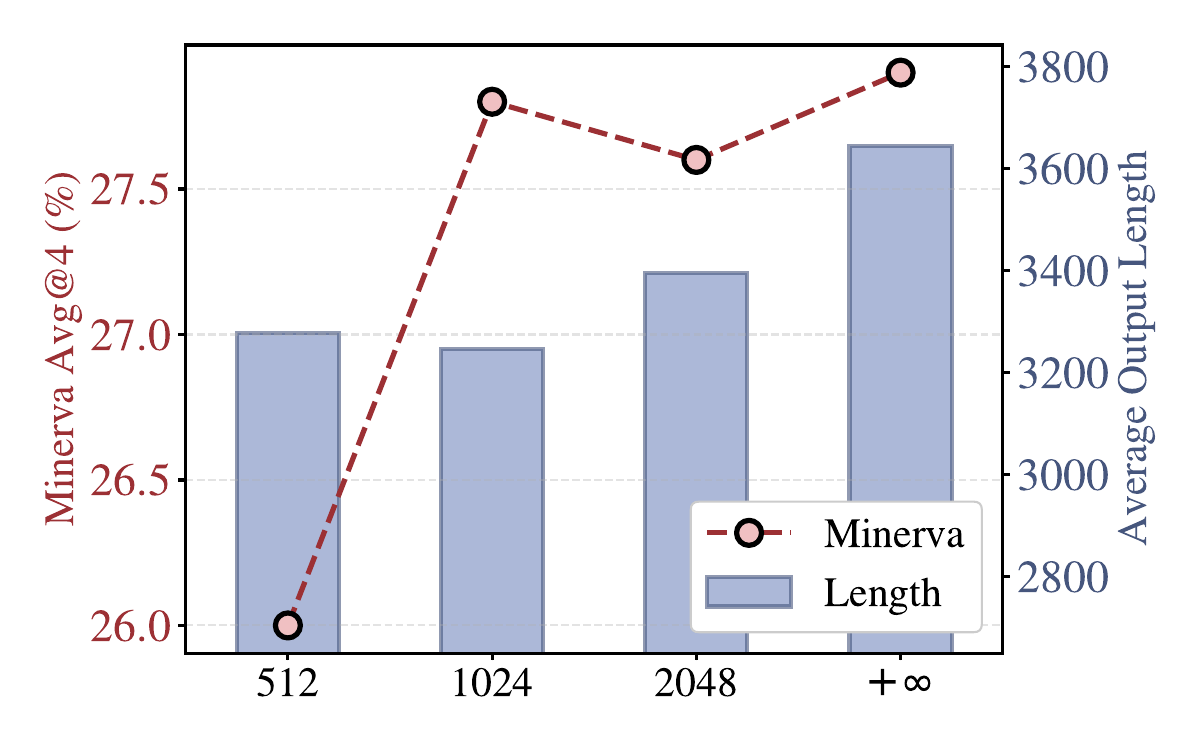}
            \subcaption{\small Analysis of $\lambda$}
            \label{fig:ab_lam}
        \end{minipage}
    \end{minipage}
    \caption{Sensitivity Analysis of $\tau$, Advantage Clipping, and $\lambda$. $\lambda=+\infty$ means RLRR w/o Length Re-ranking.}
    \vspace{-10pt}
\end{figure}

\textbf{Evaluation on Pretrained Models.} 
We evaluate the performance of diverse methods using Qwen2.5-7B~\citep{qwen2.5} as the backbone, with detailed results summarized in \Cref{tab:base_model_rl}. RLRR achieves the best performance across the majority of datasets, demonstrating particularly significant improvements on challenging benchmarks like AIME2024 and AMC. Regarding generation length, we observe minimal variation among different methods, which we attribute to the limited reasoning capacity inherent to the base model. Notably, RLRR yields strong results even when instantiated with an SRM. This finding suggests that the relative reward shaping strategy serves as a critical driver of the observed performance gains, validating its effectiveness in extracting robust training signals even from scalar models.

\textbf{Hyperparameter Sensitivity.}
We examine the impact of $\tau$ and $\lambda$. As shown in \Cref{fig:tau}, excessive $\tau$ dilutes rule-based rewards, potentially misaligning gradients with correctness and degrading performance. Introducing advantage clipping effectively mitigates this instability. As illustrated in \Cref{fig:ab_lam}, relaxing the conciseness constraint through $\lambda$ leads to longer responses. Notably, the model retains high accuracy even at reduced lengths, confirming that the length re-ranking effectively trims redundancy without compromising reasoning quality.

\section{Conclusion}

In this paper, we introduce RLRR, a framework that shifts the paradigm of group-based optimization from absolute scoring to relative preference ranking. By synthesizing intra-group comparisons through relative reward shaping, RLRR effectively mitigates gradient vanishing caused by sparse supervision and resolves the optimization instability inherent in SRMs. Extensive experiments demonstrate that RLRR yields consistent improvements across verifiable reasoning and open-ended writing tasks. These findings underscore the superiority of relative preference signals over absolute scoring for model optimization. Furthermore, we show that the Ranking RM serves as a robust listwise evaluator that naturally aligns with the comparative structure of group-based learning, delivering effective signals even with limited training data. Future work will explore more efficient mechanisms for leveraging relative rewards.

% \section*{Accessibility}

% Authors are kindly asked to make their submissions as accessible as possible
% for everyone including people with disabilities and sensory or neurological
% differences. Tips of how to achieve this and what to pay attention to will be
% provided on the conference website \url{http://icml.cc/}.

% \section*{Software and Data}

% If a paper is accepted, we strongly encourage the publication of software and
% data with the camera-ready version of the paper whenever appropriate. This can
% be done by including a URL in the camera-ready copy. However, \textbf{do not}
% include URLs that reveal your institution or identity in your submission for
% review. Instead, provide an anonymous URL or upload the material as
% ``Supplementary Material'' into the OpenReview reviewing system. Note that
% reviewers are not required to look at this material when writing their review.

% Acknowledgements should only appear in the accepted version.
% \section*{Acknowledgements}

% \textbf{Do not} include acknowledgements in the initial version of the paper
% submitted for blind review.

% If a paper is accepted, the final camera-ready version can (and usually should)
% include acknowledgements.  Such acknowledgements should be placed at the end of
% the section, in an unnumbered section that does not count towards the paper
% page limit. Typically, this will include thanks to reviewers who gave useful
% comments, to colleagues who contributed to the ideas, and to funding agencies
% and corporate sponsors that provided financial support.

\section*{Impact Statement}
This paper presents work whose goal is to advance the field of Machine Learning, particularly in the domain of reinforcement learning for Large Language Models. Our research aims to improve the reliability of models in mathematical reasoning and open-ended generation tasks. There are many potential societal consequences of advancing LLM capabilities, none of which we feel must be specifically highlighted here beyond the general discourse on AI safety and alignment.

% In the unusual situation where you want a paper to appear in the
% references without citing it in the main text, use \nocite
% \nocite{langley00}

\bibliography{rankgrpo}
\bibliographystyle{icml2026}

%%%%%%%%%%%%%%%%%%%%%%%%%%%%%%%%%%%%%%%%%%%%%%%%%%%%%%%%%%%%%%%%%%%%%%%%%%%%%%%
%%%%%%%%%%%%%%%%%%%%%%%%%%%%%%%%%%%%%%%%%%%%%%%%%%%%%%%%%%%%%%%%%%%%%%%%%%%%%%%
% APPENDIX
%%%%%%%%%%%%%%%%%%%%%%%%%%%%%%%%%%%%%%%%%%%%%%%%%%%%%%%%%%%%%%%%%%%%%%%%%%%%%%%
%%%%%%%%%%%%%%%%%%%%%%%%%%%%%%%%%%%%%%%%%%%%%%%%%%%%%%%%%%%%%%%%%%%%%%%%%%%%%%%
\newpage
\appendix
\onecolumn

\section{Training Details}

\subsection{Settings}

\label{setting}
We cap the generated output length at 8{,}192 tokens and form groups of size \(G=8\) per prompt. 
For the Ranking RM, we set the sortable subset size to \(n=4\). 
Unless otherwise noted, hyperparameters are fixed as follows:
\(\lambda = 2048\), \(\xi^+/\xi^- = \pm 10^{-3}\), 
\(\tau = 0.1\) , and sampling temperature \(T=1.0\) during data collection. 
Our method and all baselines are implemented on top of the VeRL~\citep{sheng2025hybridflow} framework.

For the reward model evaluation, we set \(n = 4\). When \(n = 2\), we repeat the process once for each sample. For \(n = 8\) or \(n = 16\), we first divide the samples into multiple groups of size 4, then select the best from each group. Afterward, we continue with the Ranking RM for a second round of selection until the optimal answer is chosen.

\subsection{Evaluation Datasets}
\label{app:datas}

We evaluate our models on seven mathematical reasoning benchmarks: Math500~\citep{hendrycksmath2021, lightman2023lets}, AIME24~\citep{AoPS:AIMEProblemsSolutions}, AIME25~\citep{AoPS:AIMEProblemsSolutions_2025}, AMC~\citep{AoPS:AMCProblemsSolutions}, Minerva Math~\citep{lewkowycz2022solving}, Gaokao~\citep{zhang2023evaluating}, and Olympiad Bench~\citep{he2024olympiadbench}, which cover a broad range of mathematical difficulty and problem types. For logical reasoning, we select two representative benchmarks: Zebra Puzzle~\citep{cheng2025revisiting}, and Ordering Puzzle~\citep{cheng2025revisiting}. These datasets are widely recognized and present diverse challenges for evaluating both mathematical and general reasoning abilities.
We configure the maximum generation length to 16,384 tokens for mathematical reasoning tasks and 32,768 tokens for logical reasoning tasks. Conversely, for open-ended writing tasks, we restrict the generation limit to 8,192 tokens and utilize the open-source Critic Model~\citep{wu2025writingbench} as the judge.

\section{Theoretical Analysis of Relative Reward Stability}
\label{app:theo_analysis}

In group-based reinforcement learning frameworks such as GRPO, policy optimization relies fundamentally on the accurate estimation of the advantage function. This estimation requires a robust baseline derived from the group statistics. However,  Scalar Reward Models (SRM) introduce instability due to the unbounded nature of their output scores. In this section, we provide a theoretical analysis demonstrating that the loss function of SRMs inherently drives rewards toward divergence, propagating high variance into the advantage estimation. We then formally prove that the proposed RLRR method mitigates this issue by imposing strict variance bounds through reward normalization.

\subsection{Gradient-Driven Divergence in Scalar Reward Models}

To understand the source of score instability, we analyze the optimization dynamics of the standard reward modeling objective. SRMs typically parameterize the preference probability using the Bradley-Terry model. Given a prompt $x$, a preferred response $y_w$, and a rejected response $y_l$, the preference probability is defined as $\sigma(\delta)$, where $\sigma$ is the sigmoid function and $\delta = r(x, y_w) - r(x, y_l)$ is the reward difference.

The training objective minimizes the negative log-likelihood over the dataset $\mathcal{D}$:

\begin{equation}
\label{eq:srm_loss}
\mathcal{L}_{\text{SRM}} = - \mathbb{E}_{\mathcal{D}} \left[ \log \sigma(\delta) \right] = \mathbb{E}_{\mathcal{D}} \left[ \log(1 + e^{-\delta}) \right]
\end{equation}

By examining the gradient of this loss function with respect to the reward difference $\delta$, we reveal the driving force behind the unbounded scores. The derivative is given by:

\begin{equation}
\frac{\partial \mathcal{L}}{\partial \delta} = \frac{-e^{-\delta}}{1 + e^{-\delta}} = \sigma(\delta) - 1
\end{equation}

Since the image of the sigmoid function for any finite input is the open interval $(0, 1)$, the gradient $\sigma(\delta) - 1$ is strictly negative for all finite $\delta$. This indicates that $\mathcal{L}_{\text{SRM}}$ is monotonically decreasing with respect to $\delta$. Consequently, the infimum of the loss is approached only as $\delta \rightarrow +\infty$.

During optimization, the gradient descent process exerts a constant pressure to maximize the gap between $y_w$ and $y_l$. In the absence of explicit regularization on the reward magnitudes, this creates a runaway effect where the absolute values of $r(x, y)$ must grow indefinitely to satisfy the objective. Furthermore, the Bradley-Terry model is invariant to translation, meaning $r(x, y)$ and $r(x, y) + C$ yield identical losses. This lack of anchoring, combined with the gradient pressure for divergence, results in a reward distribution with theoretically unbounded support. This mathematical property directly leads to the generation of extreme values and high variance during the RL training phase.

\subsection{Bias Propagation in Group-Based Advantage Estimation}

The unbounded nature of SRM outputs significantly impacts the stability of group-based estimators used in algorithms like GRPO. The advantage $A_i$ for the $i$-th sample in a group of size $G$ is computed by normalizing the reward score $s_i$ against the group statistics:

\begin{equation}
A_i = \frac{s_i - \bar{s}}{\sigma_G}
\end{equation}

Here $\bar{s}$ denotes the sample mean and $\sigma_G$ denotes the sample standard deviation. The reliability of $A_i$ depends on $\bar{s}$ being a robust estimator of the true distributional mean. However, as established in the previous subsection, the SRM tends to produce heavy-tailed or extreme score distributions.

Consider a scenario where the SRM assigns an extreme outlier score to a single sample due to model overconfidence or the divergence mechanism described above. Since the sample mean $\bar{s}$ is not robust to outliers, this single value shifts the baseline significantly. Consequently, the advantages computed for the remaining "normal" samples are skewed, often collapsing towards zero or exhibiting incorrect signs. The high dispersion of the scalar scores implies that while the ranking might be preserved, the magnitude of the advantage becomes dominated by numerical noise rather than signal quality. This variance in the baseline estimator introduces bias into the policy gradient update, thereby destabilizing the training process.

\subsection{Variance Bounds via RLRR}

The proposed Reinforcement Learning with Relative Reward (RLRR) framework resolves this instability by fundamentally altering the support of the reward distribution. Instead of relying on raw scalar outputs, RLRR enforces constraints that map rewards into a compact set.

As defined in the method section, the Pure Relative Reward maps rankings linearly to the interval $[0, 1]$, while the Hybrid Relative Reward utilizes the hyperbolic tangent function to constrain corrections within a fixed range derived from the hyperparameter $\tau$. In both cases, the reward $R$ is a random variable strictly bounded within a finite interval $[a, b]$.

We invoke Popoviciu's inequality on variances to demonstrate the theoretical benefit of this design. For any random variable $X$ with support bounded in $[a, b]$, the variance is strictly limited by:

\begin{equation}
\text{Var}(X) \leq \frac{(b - a)^2}{4}
\end{equation}

For PRR, where the range is $[0, 1]$, the variance of the reward distribution is strictly upper-bounded by $0.25$, regardless of the model confidence or training duration. Similarly, for HRR, the variance is bounded by a function of $\tau$.

By imposing this hard mathematical limit on variance, RLRR ensures that no single sample can exert an arbitrarily large influence on the group mean $\bar{s}$ or standard deviation $\sigma_G$. This guarantees that the advantage estimator remains robust even in the presence of distinct preference differences. Therefore, the transition from unbounded scalar rewards to bounded relative rewards provides a theoretical guarantee of stability for group-based policy optimization.

\section{Supplementary Results}
\label{app:full_result}

\subsection{Absolute Baseline.}

\begin{figure}[!t]
  \centering
  % \captionsetup[sub]{skip=0pt} % 子图标题紧凑

  % ---- 左侧 50%：柱状图 ----
  \begin{minipage}[b]{0.35\textwidth}
    \centering
    \begin{subfigure}[b]{\linewidth}  % 改为对齐底部
      \centering
      \includegraphics[width=\linewidth]{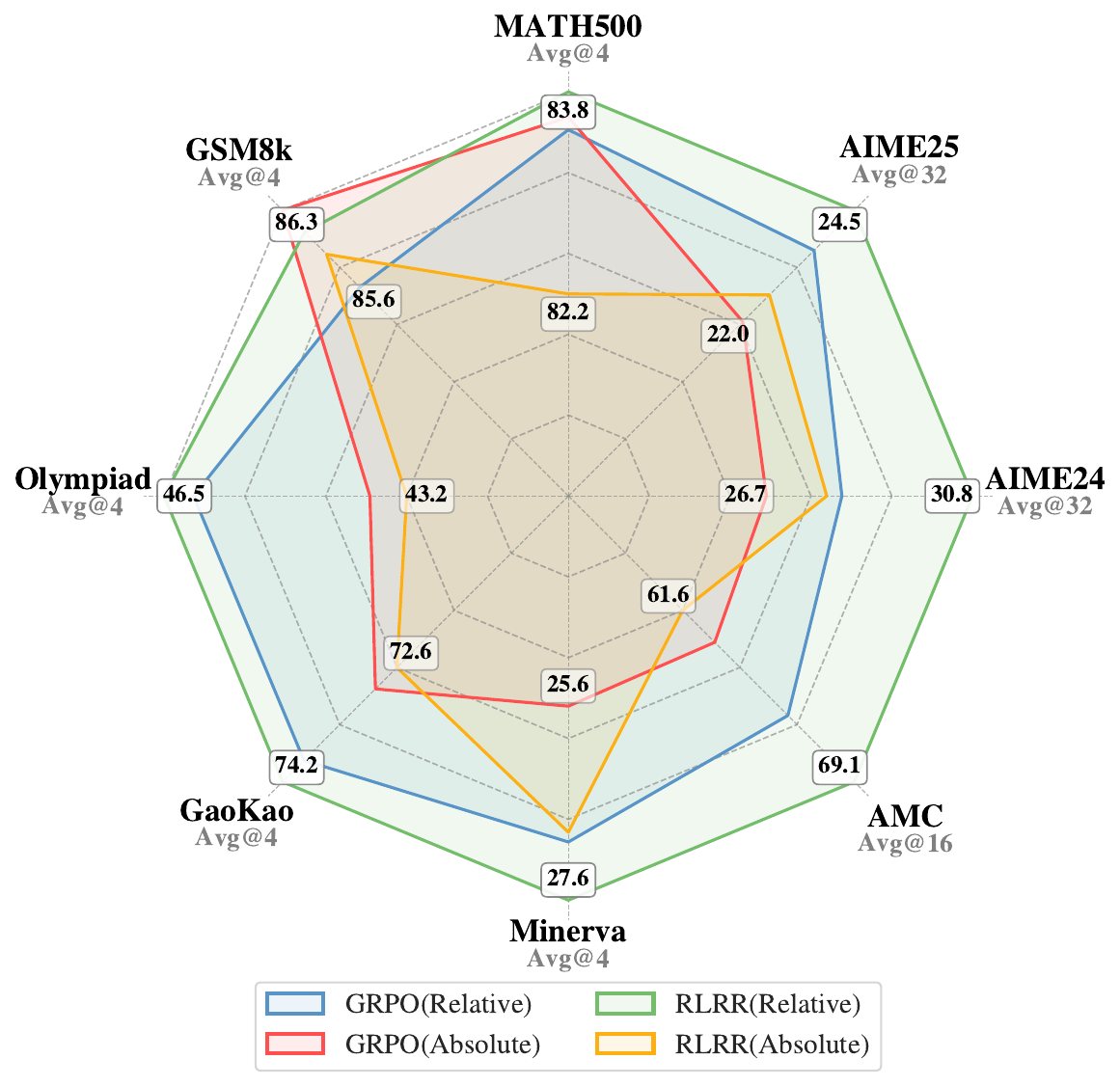}
      \subcaption{\small Overall performance comparison}
      \label{fig:ab_base_bar}
    \end{subfigure}
  \end{minipage}
  \hspace{0.005\textwidth} % 控制左边和右边的间距
  % ---- 右侧 50%：2×2 折线图 ----
  \begin{minipage}[b]{0.63\textwidth}
    \centering
    % 第一行两张
    \begin{subfigure}[b]{0.49\linewidth}  % 改为对齐底部
      \centering
      \includegraphics[width=\linewidth]{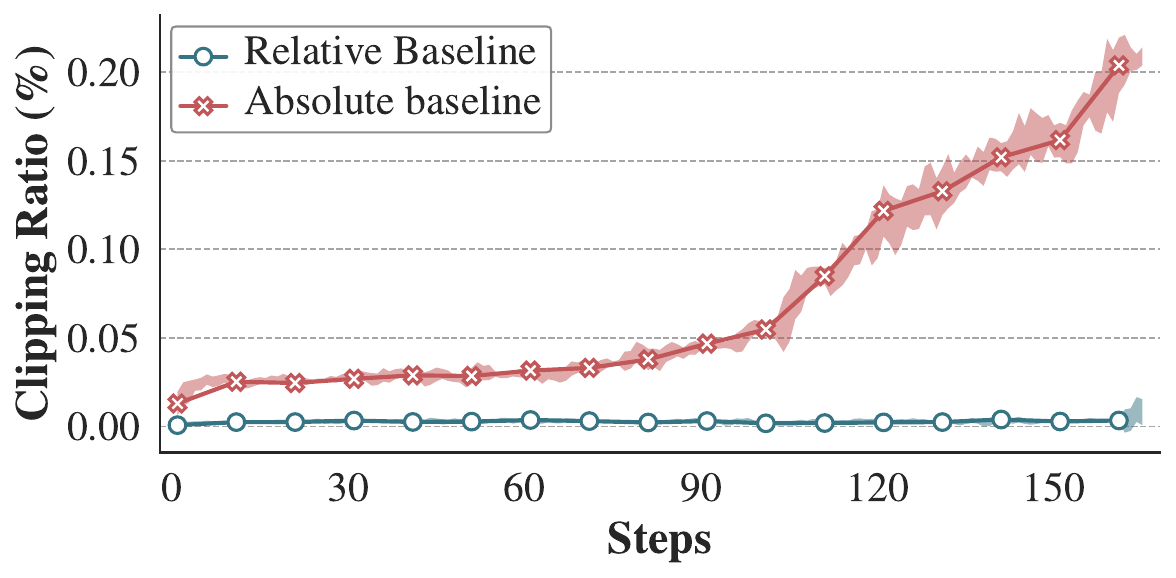}
      \subcaption{\small Training Dynamic (GRPO)}
      \label{fig:ab_dy_grpo}
    \end{subfigure}\hspace{0.005\textwidth}
    \begin{subfigure}[b]{0.49\linewidth}  % 改为对齐底部
      \centering
      \includegraphics[width=\linewidth]{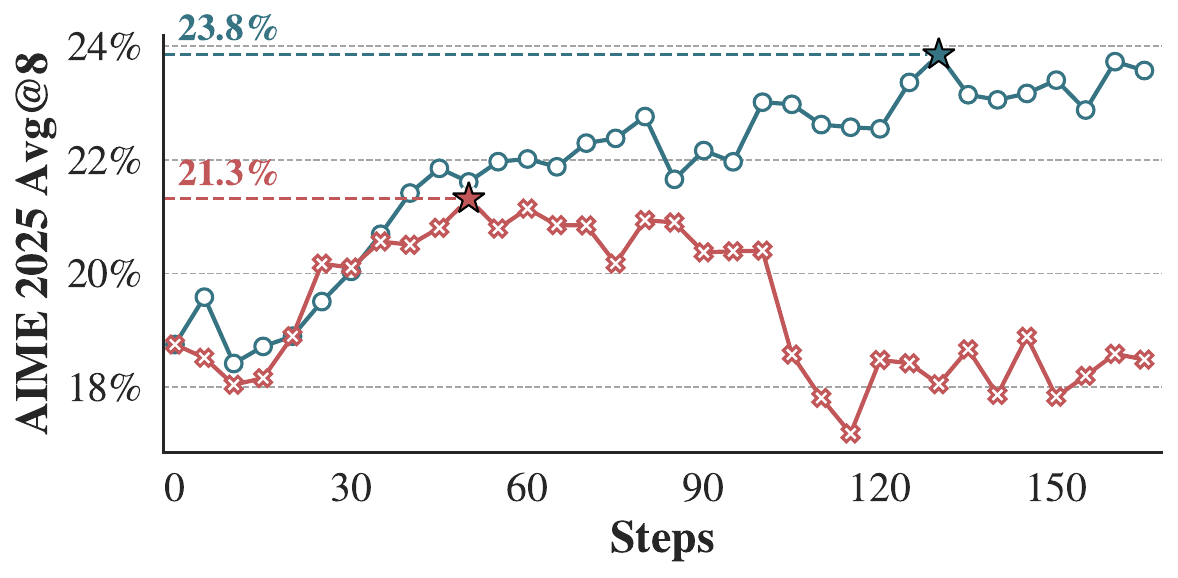}
      \subcaption{\small AIME2025 Avg@8 (GRPO)}
      \label{fig:ab_aime_grpo}
    \end{subfigure}

    \vspace{-0.2em} % 减少第二行的垂直间距

    % 第二行两张
    \begin{subfigure}[b]{0.49\linewidth}  % 改为对齐底部
      \centering
      \includegraphics[width=\linewidth]{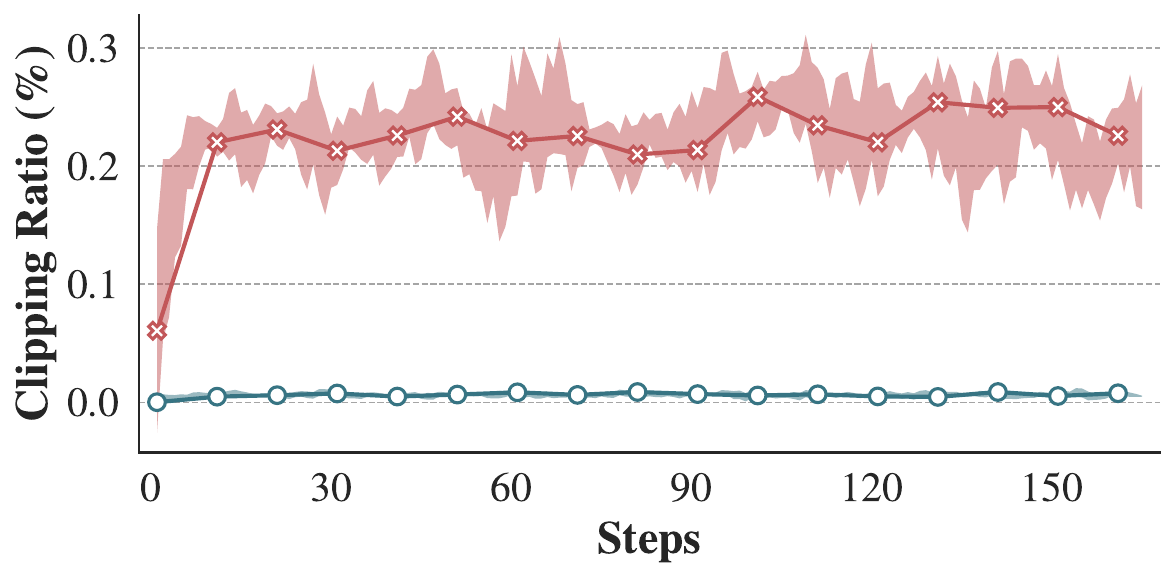}
      \subcaption{\small Training Dynamic (RLRR)}
      \label{fig:ab_dy_rank}
    \end{subfigure}\hspace{0.01\textwidth}
    \begin{subfigure}[b]{0.49\linewidth}  % 改为对齐底部
      \centering
      \includegraphics[width=\linewidth]{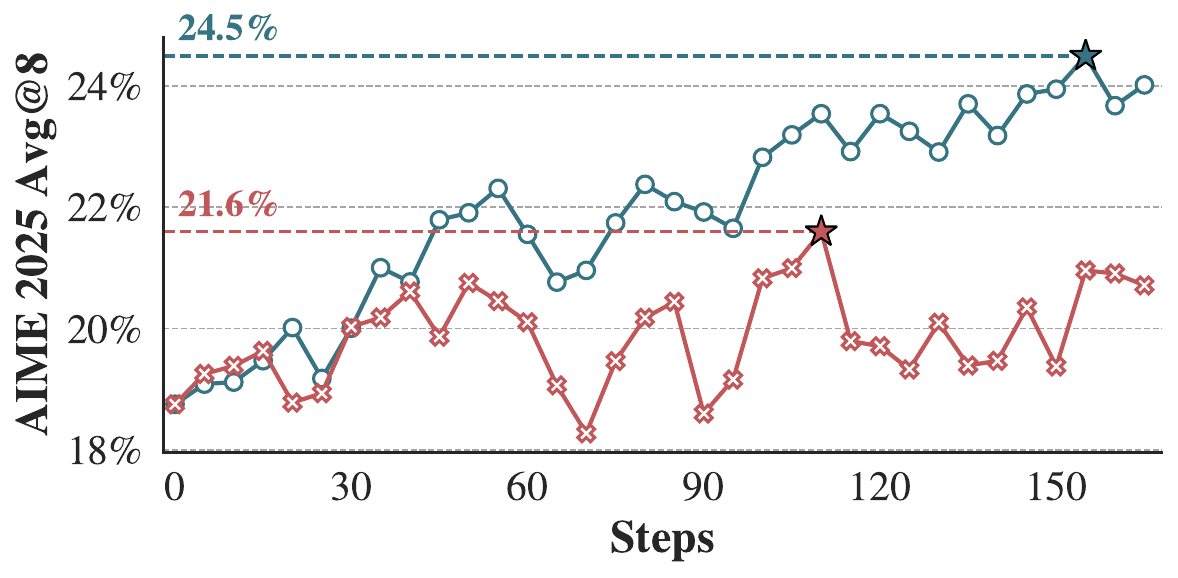}
      \subcaption{\small AIME2025 Avg@8 (RLRR)}
      \label{fig:ab_aime_rank}
    \end{subfigure}
  \end{minipage}

\caption{Effect of Absolute vs. Relative Baselines on GRPO and RLRR.}  \label{fig:baseline_ab}
\end{figure}

In our experiments, we compared the impact of using relative baselines (intra-group mean) versus absolute correctness baselines (for GRPO, a baseline of 0.5; for RLRR, a baseline of 1) on performance and stability. The results, shown in \Cref{fig:ab_base_bar}, indicate that using the absolute correctness baseline leads to a significant drop in performance.
\Cref{fig:ab_dy_grpo} and \ref{fig:ab_dy_rank} further reveal the instability introduced by the absolute baseline, particularly from the perspective of truncation rates. Additionally, \Cref{fig:ab_aime_grpo} and \ref{fig:ab_aime_rank} demonstrate a decline in accuracy during the later stages of training, highlighting the unsuitability of the absolute baseline for long-term training.

\subsection{Impact of Dataset Difficulty}
\label{app:data_level_discuss}

\begin{figure}
    \centering
    \includegraphics[width=0.75\linewidth]{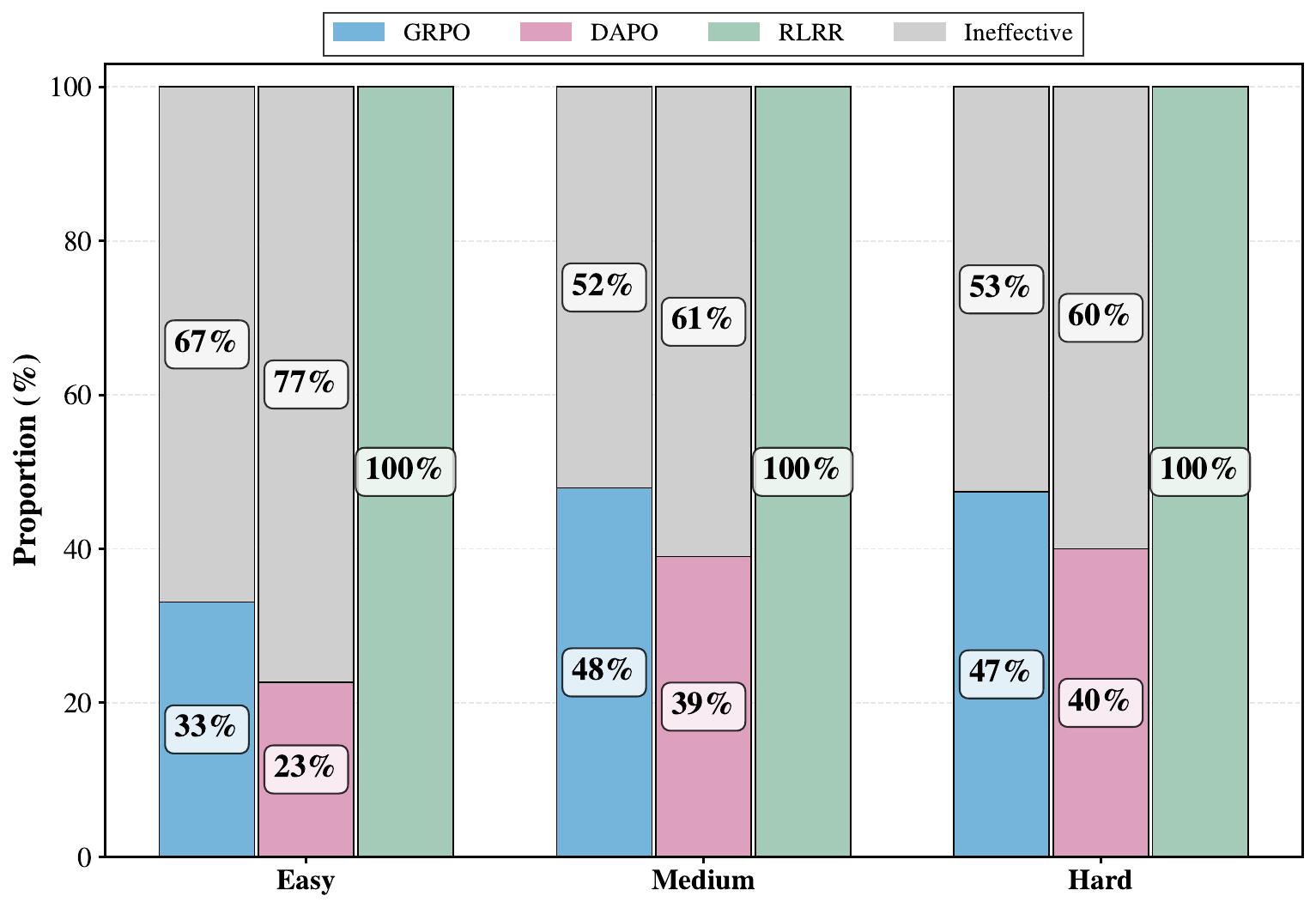}
    \caption{The proportion of effective data during the training phase for different methods.}
    \label{fig:data_level_pos}
\end{figure}

We analyze how dataset difficulty influences data efficiency and performance in~\Cref{sec:data_level_main}. 
\Cref{tab:data_level_tab} reports a detailed comparison of three methods across difficulty levels. 
Moderate difficulty yields the best gains, whereas overly easy or overly hard data diminishes further improvement. 
In GRPO, extremes of difficulty tend to degenerate into \emph{invalid prompts} that provide little learning signal and mainly act as a weak regularizer to prevent forgetting of trivial cases. In fact, on the easy subset many prompt groups are already close to unanimously correct at the beginning of training, so the GRPO effective prompt ratio starts at a relatively low level and quickly saturates. On the medium subset the GRPO effective prompt ratio decreases from about 60\% at the beginning of training to about 40\% near convergence, which is consistent with the global trend in \Cref{fig:effective_prompts}. On the hard subset many prompt groups are initially unanimously incorrect and gradually become effective as the policy improves, which compensates for prompts that later turn unanimously correct and produces an almost flat curve. Across all difficulty levels, the absolute fraction of effective prompts under GRPO remains relatively low, indicating limited utilization of the available data. 
DAPO removes such invalid prompts altogether, which avoids noise but forfeits potential information contained therein. 
By contrast, RLRR leverages \emph{all} samples by converting groupwise orderings into usable signal, thereby extracting additional knowledge even from otherwise low-value prompts. 
\Cref{fig:data_level_pos} visualizes the fraction of effective data throughout RL training: RLRR maintains \(100\%\) effective utilization at all times, substantially exceeding the other methods and corroborating its advantages in both data efficiency and final performance.

% 16k
\begin{table}[!t]
\centering
\caption{Performance comparison across datasets of varying difficulty; \textbf{bold} indicates the best result.
}
\label{tab:data_level_tab}
\resizebox{\textwidth}{!}{%
\begin{tabular}{llccccccccc}
\toprule

\multicolumn{2}{c}{\textsc{Method}} & AIME24 & AIME25 & MATH500 & GSM8k & Olympiad & GaoKao & Minerva & AMC & Avg \\
\midrule
\multirow{3}{*}{Easy} & GRPO & 28.5 & 22.8 & 82.6 & \textbf{87.0} & 44.0 & 72.3 & \textbf{28.2} & 62.7 & 53.5 \\
 & DAPO & \textbf{30.0} & \textbf{22.9} & 82.4 & 86.0 & 42.6 & 73.0 & 26.3 & 62.3 & 53.2 \\
 & RLRR & 29.7 & 22.8 & \textbf{83.2} & 86.8 & \textbf{44.1} & \textbf{74.6} & 27.4 & \textbf{64.5} & \textbf{54.1} \\
 \midrule
\multirow{3}{*}{Medium} & GRPO & 28.2 & 23.6 & 83.5 & 85.6 & 46.1 & 73.9 & 27.0 & 66.2 & 54.3 \\
 & DAPO & 29.3 & 23.2 & 82.5 & 86.0 & 42.7 & 72.8 & 25.6 & 62.0 & 53.0 \\
 & RLRR & \textbf{30.8} & \textbf{24.5} & \textbf{83.8} & \textbf{86.1} & \textbf{46.5} & \textbf{74.2} & \textbf{27.6} & \textbf{69.1} & \textbf{55.3} \\
 \midrule
\multirow{3}{*}{Hard} & GRPO & 27.9 & 23.6 & 82.9 & 85.9 & 43.4 & 74.2 & 26.3 & 65.4 & 53.7 \\
 & DAPO & 28.1 & 22.1 & 82.9 & 85.8 & 43.3 & 72.6 & \textbf{27.4} & 62.2 & 53.0 \\
 & RLRR & \textbf{28.8} & \textbf{23.8} & \textbf{83.6} & \textbf{86.2} & \textbf{45.0} & \textbf{74.9} & 26.7 & \textbf{66.4} & \textbf{54.4}\\
 \bottomrule
\end{tabular}
}
\end{table}

\subsection{Effect of Group Size on Method Performance}

% 16k
\begin{table}[t]
\centering
\caption{Performance Comparison Across Different Group Sizes}
\label{tab:group_size}
\resizebox{\textwidth}{!}{%
\begin{tabular}{llccccccccc}
\toprule
\multirow{2}{*}{$G$} & \multirow{2}{*}{\textsc{Method}} & \multicolumn{8}{c}{\textsc{Mathematical Reasoning}} &  \multirow{2}{*}{Avg} \\
\cmidrule{3-10}
 &  & AIME24 & AIME25 & MATH500 & GSM8k & Olympiad & GaoKao & Minerva & AMC &  \\
\midrule
\multirow{2}{*}{4} 
 & GRPO & 28.6 & 22.4 & 83.1 & 86.6 & 45.9 & 72.8 & 27.4 & 66.3 & 54.1 \\
 & RLRR & 31.5 & 24.1 & 84.3 & 85.1 & 45.0 & 73.8 & 26.6 & 65.4 & 54.5 \\
\cmidrule(lr){1-11}

\multirow{2}{*}{8} 
 & GRPO & 28.2 & 23.6 & 83.5 & 85.6 & 46.1 & 73.9 & 27.0 & 66.2 & 54.3 \\
 & RLRR & 30.8 & 24.5 & 83.8 & 86.1 & 46.5 & 74.2 & 27.6 & 69.1 & 55.3 \\
\cmidrule(lr){1-11}

\multirow{2}{*}{16} 
 & GRPO & 28.9 & 22.8 & 85.5 & 87.2 & 47.1 & 75.9 & 27.7 & 68.0 & 55.4 \\
 & RLRR & 30.1 & 23.2 & 85.6 & 86.8 & 46.7 & 74.9 & 28.2 & 67.8 & 55.4 \\
\bottomrule
\end{tabular}
}
\end{table}

We conducted a comprehensive analysis of both methods' performance across varying group sizes \( G \), as detailed in \Cref{tab:group_size}. The experimental results reveal that GRPO exhibits performance improvement with increasing \( G \), primarily due to reduced occurrence of invalid groups at larger group sizes. In contrast, RLRR demonstrates consistent effectiveness even at smaller \( G \) values through full utilization of all available training data. The experimental findings demonstrate that GRPO attains performance levels comparable to RLRR when operating with sufficiently large group sizes, as the increased sampling capacity enables more comprehensive data utilization.

\subsection{Effect of $\lambda$ on Method Performance}
\label{app:lambda}

We analyze the sensitivity of model performance to the parameter $\lambda$ defined in \Cref{length_rank}, as presented in \Cref{tab:lambda_analysis}. Relaxing the length constraint by increasing $\lambda$ leads to steady performance gains.  When $\lambda$ is set to a large value, the constraint becomes negligible, yielding results comparable to the configuration without length re-ranking.  On the other hand, given that the length constraint specifically targets correct responses, the model demonstrates resilience and maintains performance even under stricter constraints characterized by small $\lambda$ values.

In addition, we evaluate the robustness of the models by imposing varying maximum response length caps. While all methods exhibit some performance degradation, the approach without length re-ranking suffers the most significant drop. This empirical evidence suggests that the baseline relies heavily on inherent verbosity to achieve correctness, whereas our length re-ranking mechanism successfully biases the model towards conciseness and ensures robustness against length restrictions.

\begin{table}[htbp]
\centering
\caption{Sensitivity analysis of $\lambda$. Note that $\lambda=+\infty$ corresponds to RLRR without length re-ranking. MRL denotes the Max Response Length imposed during testing. Values in parentheses indicate the performance change relative to the 32k baseline. \textcolor{bestgreen}{Green} signifies no performance loss (optimal), \textcolor{lossgray}{gray} indicates a moderate decline, and \textcolor{worstred}{red} highlights the most significant degradation.
}
\label{tab:lambda_analysis}
\resizebox{\textwidth}{!}{%
\begin{tabular}{
    cc *{9}{S[
                table-format=2.1, 
                % table-auto-round,
                detect-weight=true, 
                mode=match,
                table-space-text-post=\hspace{1.2em}
            ]}
    c
    }
\toprule
\multirow{2}{*}{$\lambda$} & \multirow{2}{*}{MRL} & \multicolumn{9}{c}{\textsc{Mathematical Reasoning}} &  \multirow{2}{*}{Avg Len} \\
\cmidrule{3-11}
 & & {AIME24} & {AIME25} & {MATH500} & {GSM8k} & {Olympiad} & {GaoKao} & {Minerva} & {AMC} & {Avg} & \\
\midrule
% 32k
\multirow{3}{*}{512} &   32k  & 29.2 & 25.2 & 84.1 & 86.7 & 45.6 & 75.5 & 26.0 & 65.7 & 54.8 & 4773  \\
&   16k  & \chgLoss{29.0}{-0.2} & \chgLoss{25.1}{-0.1} & \chgLoss{84.0}{-0.1} & \chgLoss{86.7}{0.0} & \chgLoss{45.5}{-0.1} & \chgLoss{75.3}{-0.2} & \chgLoss{26.0}{0.0} & \chgLoss{65.7}{0.0} & \chgLoss{54.7}{-0.1} & 4441 \\
&   8k  & \chgLoss{25.6}{-3.6} & \chgLoss{23.6}{-1.6} & \chgLoss{82.8}{-1.3} & \chgLoss{86.7}{0.0} & \chgLoss{43.9}{-1.7} & \chgLossr{74.2}{-1.3} & \chgLoss{25.8}{-0.2} & \chgLoss{62.4}{-3.3} & \chgLoss{53.1}{-1.7} & 3677  \\
\midrule

\multirow{3}{*}{1024} & 32k   & 30.5 & 23.9 & 84.5 & 86.3 & 47.6 & 74.2 & 28.4 & 66.5 & 55.2 & 4698  \\
& 16k & \chgLoss{30.5}{0.0} & \chgLoss{23.8}{-0.1} & \chgLoss{84.5}{0.0} & \chgLoss{86.2}{-0.1} & \chgLoss{47.4}{-0.2} & \chgLoss{74.1}{-0.1} & \chgLossr{27.8}{-0.6} & \chgLoss{66.4}{-0.1} & \chgLoss{55.1}{-0.1} & 4409 \\
& 8k & \chgLoss{27.9}{-2.6} & \chgLoss{22.8}{-1.1} & \chgLoss{82.9}{-1.6} & \chgLoss{86.2}{-0.1} & \chgLoss{45.4}{-2.2} & \chgLoss{73.1}{-1.1} & \chgLossr{27.8}{-0.6} & \chgLoss{63.0}{-3.5} & \chgLoss{53.6}{-1.6} & 3643 \\
\midrule

\multirow{3}{*}{2048} & 32k   & 30.8 & 24.6 & 84.0 & 86.1 & 46.7 & 74.3 & 27.6 & 69.2 & 55.4 & 4972 \\
& 16k & \chgLoss{30.8}{0.0} & \chgLoss{24.5}{-0.1} & \chgLoss{83.8}{-0.2} & \chgLoss{86.1}{0.0} & \chgLoss{46.5}{-0.2} & \chgLoss{74.2}{-0.1} & \chgLoss{27.6}{0.0} & \chgLoss{69.1}{-0.1} & \chgLoss{55.3}{-0.1} & 4518 \\
& 8k & \chgLoss{27.2}{-3.6} & \chgLoss{23.6}{-1.0} & \chgLoss{82.7}{-1.3} & \chgLoss{86.1}{0.0} & \chgLossr{42.7}{-4.0} & \chgLoss{73.2}{-1.1} & \chgLoss{27.1}{-0.5} & \chgLoss{65.6}{-3.6} & \chgLossr{53.4}{-2.0} & 3676 \\

\midrule

\multirow{3}{*}{$+\infty$} & 32k  & 31.3 & 24.8 & 85.1 & 86.9 & 47.7 & 73.8 & 28.0 & 66.9 & 55.6 & 5255  \\ 
& 16k & \chgLoss{31.3}{0.0} & \chgLoss{24.7}{-0.1} & \chgLoss{85.0}{-0.1} & \chgLoss{86.9}{0.0} & \chgLoss{47.2}{-0.5} & \chgLoss{73.8}{0.0} & \chgLoss{27.9}{-0.1} & \chgLoss{66.8}{-0.1} & \chgLoss{55.4}{-0.2} & 4743 \\
&8k& \chgLossr{27.5}{-3.8} & \chgLossr{23.0}{-1.8} & \chgLossr{82.8}{-2.3} & \chgLossr{86.5}{-0.4} & \chgLoss{45.4}{-2.3} & \chgLoss{72.9}{-0.9} & \chgLossr{27.4}{-0.6} & \chgLossr{63.0}{-3.9} & \chgLossr{53.6}{-2.0} & 3938 \\

\bottomrule
\end{tabular}
}
\end{table}

\subsection{Computational Efficiency Analysis}
\label{app:efficiency_analysis}

While our approach incurs a marginal increase in computational cost compared to rule-based baselines due to the necessity of model inference, it yields a substantial improvement in data utilization. Rule-based methods frequently discard sample groups that exhibit zero reward variance, resulting in a significantly lower effective data ratio. In contrast, RLRR exploits all generated rollouts by introducing fine-grained relative rankings, ensuring maximum training efficiency per step.

Furthermore, our method demonstrates superior inference efficiency compared to the SRM. While SRMs operate under a pointwise paradigm that requires evaluating each response individually, the Ranking RM processes $n$ samples simultaneously within a single inference pass. This batch processing capability significantly reduces the overhead associated with reward calculation. Consequently, the Ranking RM exhibits no significant disparity in computational overhead compared to the SRM, and in fact, achieves a slight reduction in resource usage. Table~\ref{tab:resource_usage} presents an empirical estimation of the resources required for training 100 steps on A100 GPUs.

\begin{table}[h]
    \centering
    \caption{Comparison of data efficiency and computational resource usage estimated over 100 training steps on A100 GPUs.}
    \label{tab:resource_usage}
    \begin{tabular}{lcc}
        \toprule
        \textbf{Method} & \textbf{Effective Data Ratio} & \textbf{Training Resources (GPU Hours)} \\
        \midrule
        Rule-based Verifier & 48\% & 72.8 \\
        Scalar Reward Model & 100\% & 85.5 \\
        \rowcolor{gray!20}
        Ranking Reward Model (Ours) & \textbf{100\%} & \textbf{79.1} \\
        \bottomrule
    \end{tabular}
\end{table}

\begin{table}[!t]
\centering
\caption{Ablation study on the effectiveness of ranking mechanisms.
\textbf{bold} indicates the best result.
}
\label{tab:rand_ranking_ablation}
\begin{tabular}{lccccc}
\toprule
\textbf{Method} & \textbf{AIME24} & \textbf{AIME25} & \textbf{Minerva} & \textbf{AMC} & \textbf{Avg} \\ 
\midrule
GRPO & 28.2 & 23.6 & 27.0 & 66.2 & 36.3 \\
Random Ranking w/ Correctness Re-ranking & 29.9 & 24.9 & 27.2 & 68.0 & 37.5 \\
\textbf{Ranking Reward Model (Ours)} & \textbf{30.8} & \textbf{24.5} & \textbf{27.6} & \textbf{69.1} & \textbf{38.0} \\ 
\bottomrule
\end{tabular}
\end{table}

\subsection{Analysis of Ranking Reliability}

We investigate the necessity of model-based relative ranking by introducing a random baseline that also incorporates correctness-aware hierarchy. In the "Random Ranking w/ Correctness Re-ranking" setting, the strict prioritization of correct samples over incorrect ones is maintained, but the relative order among responses sharing the same correctness label is randomized. This allows us to isolate the specific contribution of the Ranking RM from the benefits provided by the correctness constraint.

As shown in Table~\ref{tab:rand_ranking_ablation}, the introduction of correctness re-ranking yields an incremental improvement over the standard GRPO baseline, with the average accuracy rising from 36.3\% to 37.5\%. This result confirms that enforcing a hard constraint on correctness provides a stable foundation for the advantage signal. Furthermore, the Ranking RM achieves the best performance with an average accuracy of 38.0\%. The gain observed over the random baseline demonstrates that the Ranking RM effectively captures nuanced quality differences within the same correctness category, providing a more reliable optimization signal than mere random ordering.

\subsection{Case Study}
\label{app:case}

\begin{figure}[!t]
    \centering
    \includegraphics[width=1\linewidth]{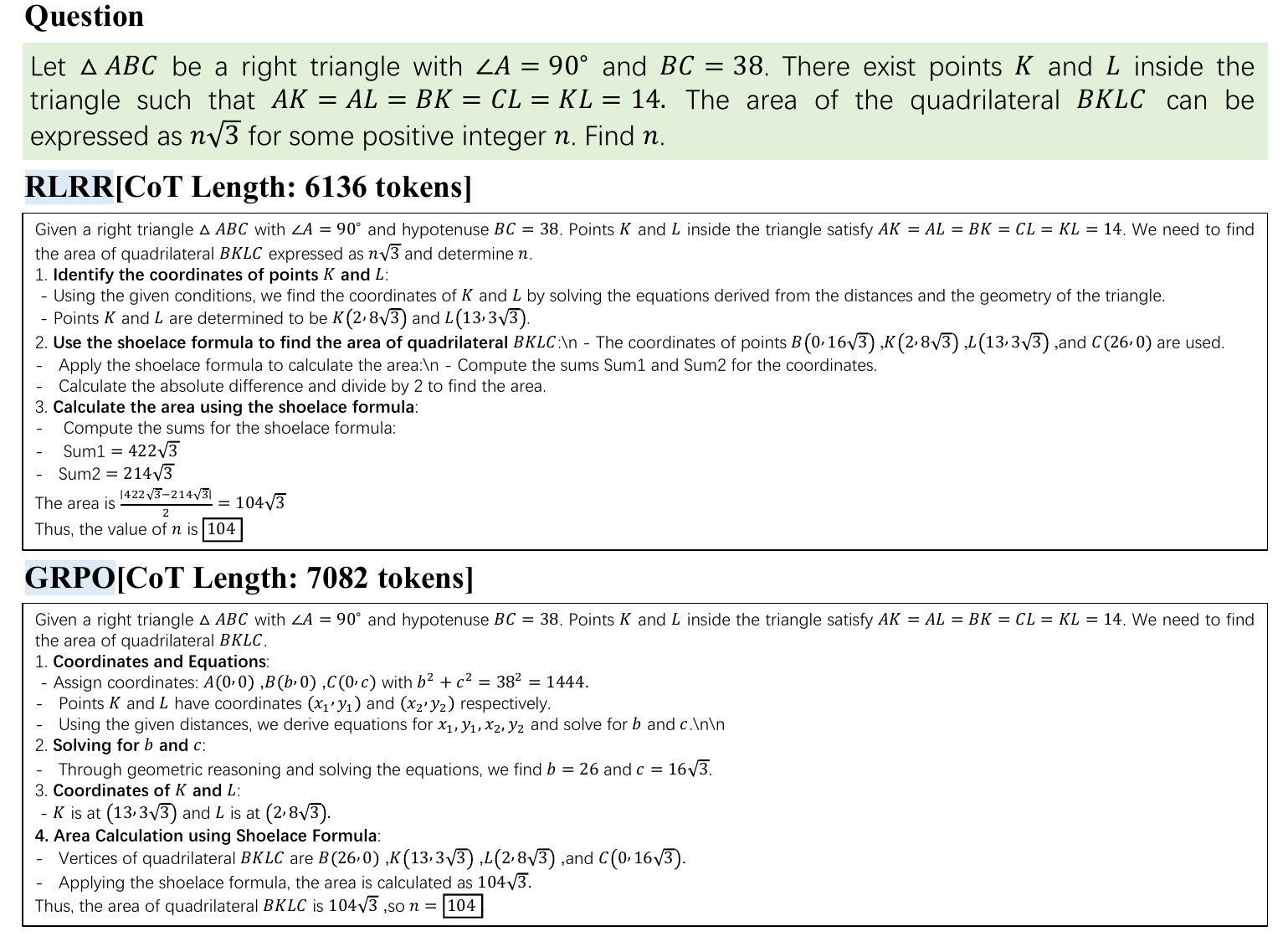}
    \caption{Performance on mathematical reasoning tasks, highlighting fewer reasoning tokens and clearer solution paths.}
    \label{fig:math_case}
\end{figure}

We present the performance on mathematical data in \Cref{fig:math_case}. Since our method encourages exploring the optimal reasoning path while ensuring correctness, the number of reasoning tokens is relatively low, and the solution approach is clearer.

% \section{You \emph{can} have an appendix here.}

% You can have as much text here as you want. The main body must be at most $8$
% pages long. For the final version, one more page can be added. If you want, you
% can use an appendix like this one.

% The $\mathtt{\backslash onecolumn}$ command above can be kept in place if you
% prefer a one-column appendix, or can be removed if you prefer a two-column
% appendix.  Apart from this possible change, the style (font size, spacing,
% margins, page numbering, etc.) should be kept the same as the main body.
%%%%%%%%%%%%%%%%%%%%%%%%%%%%%%%%%%%%%%%%%%%%%%%%%%%%%%%%%%%%%%%%%%%%%%%%%%%%%%%
%%%%%%%%%%%%%%%%%%%%%%%%%%%%%%%%%%%%%%%%%%%%%%%%%%%%%%%%%%%%%%%%%%%%%%%%%%%%%%%

\end{document}